\documentclass[journal,twoside,web]{ieeecolor}

\usepackage{etoolbox}
\makeatletter
\@ifundefined{color@begingroup}%
  {\let\color@begingroup\relax
   \let\color@endgroup\relax}{}%
\def\fix@ieeecolor@hbox#1{%
  \hbox{\color@begingroup#1\color@endgroup}}
\patchcmd\@makecaption{\hbox}{\fix@ieeecolor@hbox}{}{\FAILED}
\patchcmd\@makecaption{\hbox}{\fix@ieeecolor@hbox}{}{\FAILED}

\usepackage{jsen}
\usepackage{textcomp}
\usepackage{ifpdf}
\usepackage{cite}
\usepackage{graphicx}
\usepackage{amsmath,amssymb,amsfonts}
\usepackage{algorithm}
\usepackage{algpseudocode}
\usepackage{array}
\usepackage{url}
\usepackage{booktabs}
\usepackage{tabularx}
\usepackage{multirow}
\usepackage{wrapfig}
\usepackage{makecell}

\def\BibTeX{{\rm B\kern-.05em{\sc i\kern-.025em b}\kern-.08em
    T\kern-.1667em\lower.7ex\hbox{E}\kern-.125emX}}
\definecolor{abstractbg}{rgb}{0.89804,0.94510,0.83137}
\begin{document}
%
\title{A Light-eight Deep Human Activity Recognition Algorithm Using Multiknowledge Distillation}
%
%
%
%

\author{Runze~Chen,~\IEEEmembership{Graduate~Student~Member,~IEEE,}
	Haiyong~Luo,~\IEEEmembership{Member,~IEEE,}
	Fang~Zhao,~\IEEEmembership{Member,~IEEE,}
	Xuechun~Meng,
	Zhiqing~Xie,
	and~Yida~Zhu
	\thanks{(Corresponding authors: Haiyong Luo and Fang Zhao.)
	}}

\IEEEtitleabstractindextext{
\fcolorbox{abstractbg}{abstractbg}{%
\begin{minipage}{\textwidth}%
\begin{wrapfigure}[16]{r}{3in}%
\includegraphics[width=3in]{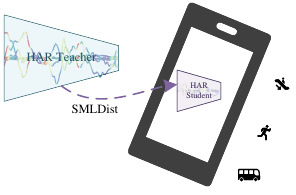}%
\end{wrapfigure}%
\begin{abstract}
Human Activity Recognition (HAR) is crucial in fields such as human-computer interaction, motion estimation, and intelligent transportation. Yet, attaining high accuracy in HAR, especially in scenarios limited by computing resources, poses a considerable challenge. This paper presents SMLDist (Stage-Memory-Logits Distillation), a framework designed to build highly customizable HAR models that achieve optimal performance under various resource constraints. SMLDist prioritizes frequency-related features in its distillation process to bolster HAR classification robustness. We also introduce an auto-search mechanism within heterogeneous classifiers to boost performance further. Our evaluation addresses the challenges of generalizing across users, sensor placements, and recognizing a wide array of activity modes. Models crafted with SMLDist, leveraging a teacher-based approach that achieves a 40\%-50\% reduction in operational expenditure, surpass the performance of existing state-of-the-art architectures. When assessing computational costs and energy consumption on the Jetson Xavier AGX platform, SMLDist-based models show strong economic and environmental sustainability advantages. Our results indicate that SMLDist effectively alleviates the performance degradation typically associated with limited computational resources, underscoring its significant theoretical and practical contributions to the field of HAR.
\end{abstract}
\end{minipage}}}

\maketitle



%

\section{Introduction}\label{sec:introduction}

%
%
%
%
\IEEEPARstart{I}{nertial} Measurement Unit (IMU)-based Human Activity Recognition (HAR) plays a pivotal role in numerous mobile sensing applications, offering critical insights for tasks such as motion estimation, intelligent transportation \cite{10.1145/3410530.3414349}, and human-computer interaction \cite{elderly-care, MHEALTH-2, mobile-games} across various domains like motion modes, traffic modes, and other related fields. HAR represents a significant application of IMUs in smart devices, with its theoretical innovations holding potential for transfer to other IMU-based tasks. As a result, researchers extensively conduct studies on the design of IMU-based HAR algorithms. Many traditional HAR algorithms primarily focus on extracting manually designed statistical features \cite{10.1145/2499621}. However, HAR based on manual feature engineering is limited by its lack of effective knowledge filtration and high dependency on specific datasets and activity types. These constraints significantly hinder its adaptability in scenarios involving deployment for new users, diverse wearable configurations, and finer-grained downstream tasks, indicating the necessity for more dynamic and flexible approaches to address these challenges. In recent years, researchers have aimed to develop smarter, more accurate, and easier-to-design HAR algorithms to address these limitations. To comprehensively extract environmental context features and user motion status features from Inertial Measurement Unit (IMU) sensor data, HAR based on deep learning methods has emerged as an effective approach.

Since the advent of deep learning methods, numerous studies focus on designing sophisticated model architectures to efficiently extract relevant feature knowledge for HAR. Initially, researchers utilize multi-layer convolutional neural network (CNN) filters to thoroughly extract activity features over short durations and employ recurrent neural network (RNN) \cite{deepconvlstm,sparsesense,indrnn-acm,10.1145/3341162.3345571,10.1145/3410530.3414349} architectures to enhance the extraction of both local and global features due to their temporal characteristics. With the progression of deep learning, mechanisms such as attention \cite{attention-is-all-you-need,attnsense,10.1145/3550331} and graph neural networks (GNNs) \cite{globalfusion,10172911,10041829} now facilitate information fusion among different sensors and wearable positions, offering a more integrated approach to understanding activity patterns. Innovative designs in model architectures enhance the capability to extract HAR-specific features. However, we identify that the training mechanisms of HAR models still have room for improvement. Complex units combine to form meticulously designed model architectures, yet they lack acceleration support across a broader range of hardware and software environments. Complex and highly coupled model structures essentially eliminate the ability to tailor models further according to deployment environments.


Our aim is to enhance the performance consistency of IMU-based HAR models across various hardware and software environments as well as application scenarios, enabling controllable model customization in computation-constrained settings while minimizing performance degradation. This approach not only reduces the complexity of deploying models on embedded platforms like smartphones, improving user experience, but also significantly lowers the energy consumption of models deployed on cloud platforms, contributing to environmental protection and creating cost advantages. Therefore, we must develop a series of processes to customize simple models, such as MobileNet V3 \cite{mobilenetv3}, thereby maximizing classification metrics to surpass current state-of-the-art models. At the same time, we aim to thoroughly investigate the lower bounds of computational complexity, space complexity, and energy consumption under the most optimal classification performance.

\begin{figure*}[thbp]
    \centering
    \includegraphics[width=\textwidth]{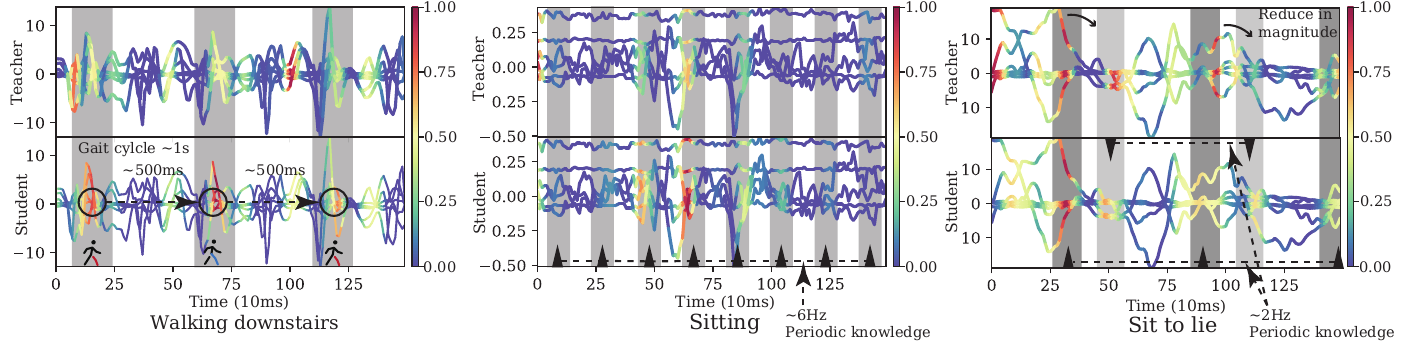}
    \caption{Implicit frequency domain knowledge in HAR tasks. The color of the raw sensor signal demonstrates the Class Activation Map (CAM) of SMLDist models \cite{DBLP:conf/cvpr/ZhouKLOT16}, where the model focuses on signals with warmer colors and ignores signals with colder colors. The gray bar represents the inherent periodicity of human activities, such as walking periodicity.}
    \label{fig:cam}
\end{figure*}

We draw inspiration from knowledge distillation (KD) techniques, which motivate us to distill HAR-specific knowledge into student models. The original KD method, proposed by Hinton et al. \cite{hinton2015distilling}, distills the teacher model's response as a soft probabilistic target, leading to increased interest in knowledge distillation among researchers. To guide student models with alternative forms of knowledge, various knowledge distillation methods have been developed, including feature-based approaches \cite{similarity-preserving,9007474} and structure-based methods \cite{DBLP:conf/cvpr/LiuJTVZGW20,Dong_2023_CVPR}. Unlike data from other modalities, IMU data exhibit distinct characteristics. As illustrated in Figure \ref{fig:cam}, users' movements display significant periodicity, and the time spans of activity cycles are broad, often extending beyond the limited receptive fields of convolutional models. Therefore, we aim to pay closer attention to the frequency domain knowledge in the features during knowledge distillation. This approach will guide the student models to avoid overfitting to local features during the learning process and to focus more on effective activity knowledge. For different application scenarios, the model's classifier also has significant customization potential. During the training process, the classifier plays a crucial role in determining the final probability outcomes. Engaging a trained classifier in the training process can introduce high-level knowledge from the teacher model's classifier, thereby guiding the training of the backbone model. Furthermore, through the teacher's training process, we can infer the adaptability of different classifier structures to specific tasks using learnable weights. With these considerations, we strive to incorporate as much knowledge as possible, such as time-frequency insights and classifier structure knowledge, obtained from initial training. This strategy significantly enhances the customizability of IMU-based models, striving for efficient and accurate HAR applications.

Drawing on the aforementioned ideas, we propose an innovative multi-level distillation pipeline, referred to as Stage-Memory-Logits Distillation (SMLDist), for IMU-based HAR modeling. SMLDist facilitates comprehensive knowledge distillation to establish robust deep HAR algorithms by distilling knowledge at three distinct levels: stage, memory, and logits. The stage knowledge encompasses the teacher model's comprehension of motion patterns and periodicity. Memory knowledge incorporates the teacher model's structural understanding of diverse classifiers, including their associated parameters. The student model effectively leverages these heterogeneous classifiers from the teacher model to efficiently assimilate the distilled knowledge within a few training epochs. By considering diverse forms of knowledge that play a critical role in training deep HAR models, SMLDist significantly improves the recognition accuracy of traditional deep HAR models without introducing additional computational overhead. To summarize, SMLDist contributes to the following advancements:

\begin{itemize}
    \item SMLDist focuses on the prominent time-frequency characteristics in HAR by leveraging teacher models to alleviate the long-term periodic patterns extracted from user activities, thereby forming stage knowledge. Through time-domain and frequency-domain distillation, it significantly mitigates the performance degradation of lightweight models when scaled down.
    \item We devise a straightforward mechanism in SMLDist that employs learnable weights tailored for heterogeneous classifiers, enabling automated customization of classifier structures for various tasks. Additionally, through parameter distillation, we imbue the student model with the classifier memory of the teacher model (parameters, structures, etc.), guiding the student model accordingly. With this approach, SMLDist achieves customization of different components such as the backbone model and classifier, presenting a more comprehensive solution for customizing HAR models in resource-constrained environments.
    \item We establish a comprehensive benchmark to thoroughly assess the performance of different model architectures across challenging scenarios such as generalization of users, generalization of sensor displacement, and diverse activity recognition. We extensively compare the performance of SMLDist with various knowledge distillation model structures in the context of HAR applications. Our findings demonstrate that, in resource-constrained environments, SMLDist effectively mitigates the performance degradation of HAR models as their scale decreases. SMLDist maintains competitive advantages in terms of computational complexity, space complexity, and energy consumption while ensuring optimal recognition performance.
\end{itemize}

The structure of this paper is as follows: Section \ref{sec:related-work} provides an overview of related works on HAR and knowledge distillation. In Section \ref{sec:method}, we present the methods used in SMLDist. In Section \ref{section:experimental-evaluation}, we conduct experiments to assess the effectiveness and performance of these methods within the SMLDist framework. Finally, Section \ref{section:conclusions-and-future-work} concludes the paper and discusses future directions.

\section{Related Work}
\label{sec:related-work}

\subsection{Human Activity Recognition}

HAR involves collecting data from external \cite{9795869} and wearable devices \cite{9389739} to analyze and identify the user's current activity state, such as daily movement \cite{realworld-har}, transportation modes \cite{htc-tm,shl-dataset-analysis}, and work status \cite{Skoda-1,Skoda-2}. HAR systems built with external devices like cameras \cite{9834306}, RGB-D \cite{10191300}, and radar \cite{9861256} offer intuitive operation and high recognition accuracy but require static deployment of sensing equipment in application scenarios. These systems also demand significant computational resources for data processing, making them suitable for fixed settings where high precision in activity recognition is crucial. On the other hand, IMU-based activity recognition stands out for its real-time capabilities, independence from specific environments, and lower energy consumption for computation \cite{app8030418}, making it well-suited for recognizing a wide range of daily activities.

Although HAR systems of different modalities exhibit significant differences, a common characteristic is their need to perceive changes in a user's state over time, which typically presents distinct time-series features. We believe that mining both the time-domain and frequency-domain knowledge significantly aids in enhancing the training quality of HAR systems\cite{indrnn-acm,7742919}. Considering the resilience of wearable devices' IMU-based HAR methods to environmental factors, their clear temporal features, and extensive coverage of user scenarios, we decide to develop a comprehensive suite of customized knowledge distillation techniques for IMU-based HAR tasks \cite{9389739}. This approach not only tailors to the specific needs of IMU-based HAR but also offers valuable insights for mining user activity knowledge from other modalities.

\subsection{Deep Learning for IMU-based HAR}

The development of deep learning technologies offers numerous approaches for HAR, where deep convolutional filters expand the receptive field on original IMU measurements. This expansion facilitates learning from local activity patterns. SparseSense \cite{sparsesense} employs multi-dimensional MLPs (Multilayer Perceptrons) to integrate these local patterns globally and generate probability predictions. RNNs and their variants introduce more ways to integrate local temporal features for HAR, using structures like LSTM (Long Short-Term Memory) \cite{deepconvlstm}, GRU (Gated Recurrent Unit) \cite{attnsense}, and IndRNN \cite{indrnn-acm} to extract global temporal knowledge over larger periods.
The introduction of RNNs enhances the temporal perception capabilities of deep learning models but significantly reduces computational throughput and implicitly increases energy consumption \cite{10.1145/3243176.3243184}. 

In HAR tasks, user activity exhibits significant frequency domain characteristics, and the analysis of frequency domain knowledge has always received considerable attention in HAR technologies beyond IMU \cite{LI2021102492,7742919}. IndRNN \cite{indrnn-acm} incorporates both time-domain and frequency-domain features as sequential inputs, highlighting the role of frequency-domain features. This approach introduces numerous time-frequency transformations into the model architecture, leading to potential increases in computational costs. Recently, researchers increasingly focus on fusing information from multiple sensors, proposing new approaches through models that leverage attention mechanisms \cite{attnsense} and GNNs \cite{globalfusion,10.1145/3550331,10041829}. EmbraceNet \cite{embracenet, embracenet-source} extracts correlated information between different sensor modalities via embracement layers for modality fusion. These approaches model the correlations between sensors as graph structures and employ graph attention for fusion. Additional sensor fusion components introduce more constraints to these HAR model structures, preventing customization and trimming of the model in many resource-constrained scenarios.

We revisit HAR methods from the perspective of training approaches, specifically addressing the limitations introduced by the complex components mentioned above. Our goal is to incorporate constraints on the periodicity or frequency domain of user activity patterns through optimized training objectives, thereby achieving effective HAR model training methods.

\subsection{Knowledge Distillation}

Knowledge distillation \cite{hinton2015distilling} is a method that enhances training quality by introducing a teacher model, featuring significant approaches such as distillation based on feature knowledge \cite{feature-normalized-kd,DBLP:conf/cvpr/LiL0Z20,DBLP:conf/cvpr/LiPYWLLC20}, structural knowledge \cite{DBLP:conf/cvpr/LiuJTVZGW20,Dong_2023_CVPR}, and response knowledge \cite{hinton2015distilling,conditional-teacher-student-learning}. Researchers refine domain knowledge through teacher models, efficiently customizing and compressing deep learning models. 

The earliest form of knowledge distillation, known as response-based knowledge distillation, uses the teacher model's responses as soft labels for auxiliary constraints in training the student model \cite{hinton2015distilling}. LHAR (Lightweight Human Activity Recognition) \cite{10195168} builds a teacher model through model integration and distills HAR-related knowledge into the student model using response-based knowledge distillation and data augmentation. Introducing conditional control into this form of knowledge distillation reduces the negative impact that the teacher model's predictions of negative samples have on the student \cite{conditional-teacher-student-learning}.

Guiding the student model with the model's responses provides implicit soft relationships between categories but offers limited knowledge to the student model. By incorporating features from the penultimate layer as a regularization term into response-based knowledge distillation, \cite{feature-normalized-kd} reduces the impact of noise in the student model's predictions. Romero et al. \cite{fitnets} use hint feature distillation to convert wide and deep models into thinner and deeper ones. Neuron selectivity transfer \cite{DBLP:journals/corr/HuangW17a} uses MMD (Maximum Mean Discrepancy) loss \cite{mmd} to metric the distance in Gaussian space between the hint features in the teacher and student models. Similarity-preserving KD \cite{similarity-preserving} distills the similarity matrix of intermediate neurons from the teacher to the student. Factor transfer \cite{DBLP:conf/nips/KimPK18} designs an encoder-decoder-styled module to extract the factor of the teacher model's intermediate features and uses an encoder for the student model to mimic the factor from the teacher model. However, regarding the temporal characteristics of HAR, suitable capabilities to represent periodic knowledge are still missing. We focus on how the periodic features of user activities can undergo effective distillation through representation in the frequency domain.

Knowledge distillation can significantly guide the tuning of model architectures. KD allows the distillation of structural knowledge into the neural search space, enhancing the search efficiency for neural blocks \cite{DBLP:conf/cvpr/LiPYWLLC20}. Additionally, aligning pretrained classifiers, decoders, and other components with the backbone model can effectively boost the performance of algorithms based on lightweight backbone models in specific tasks \cite{DBLP:journals/corr/abs-2306-14289}. Drawing on these ideas, we introduce the existing memory knowledge of classifiers into the optimization of lightweight backbone models. 

We integrate feature knowledge, structural knowledge, and response knowledge into the training process of HAR models, incorporating the abundant periodic characteristics present in HAR tasks, model structural properties suitable for downstream tasks, and soft relational features between categories into the construction of HAR models. This approach forms the SMList method, which aids in customizing HAR models with optimal classification performance, computational efficiency, and energy friendliness across various hardware and software deployment scenarios.

\section{The Pipeline of SMLDist}
\label{sec:method}

\begin{figure*}
	\centering
	\includegraphics[width=\textwidth]{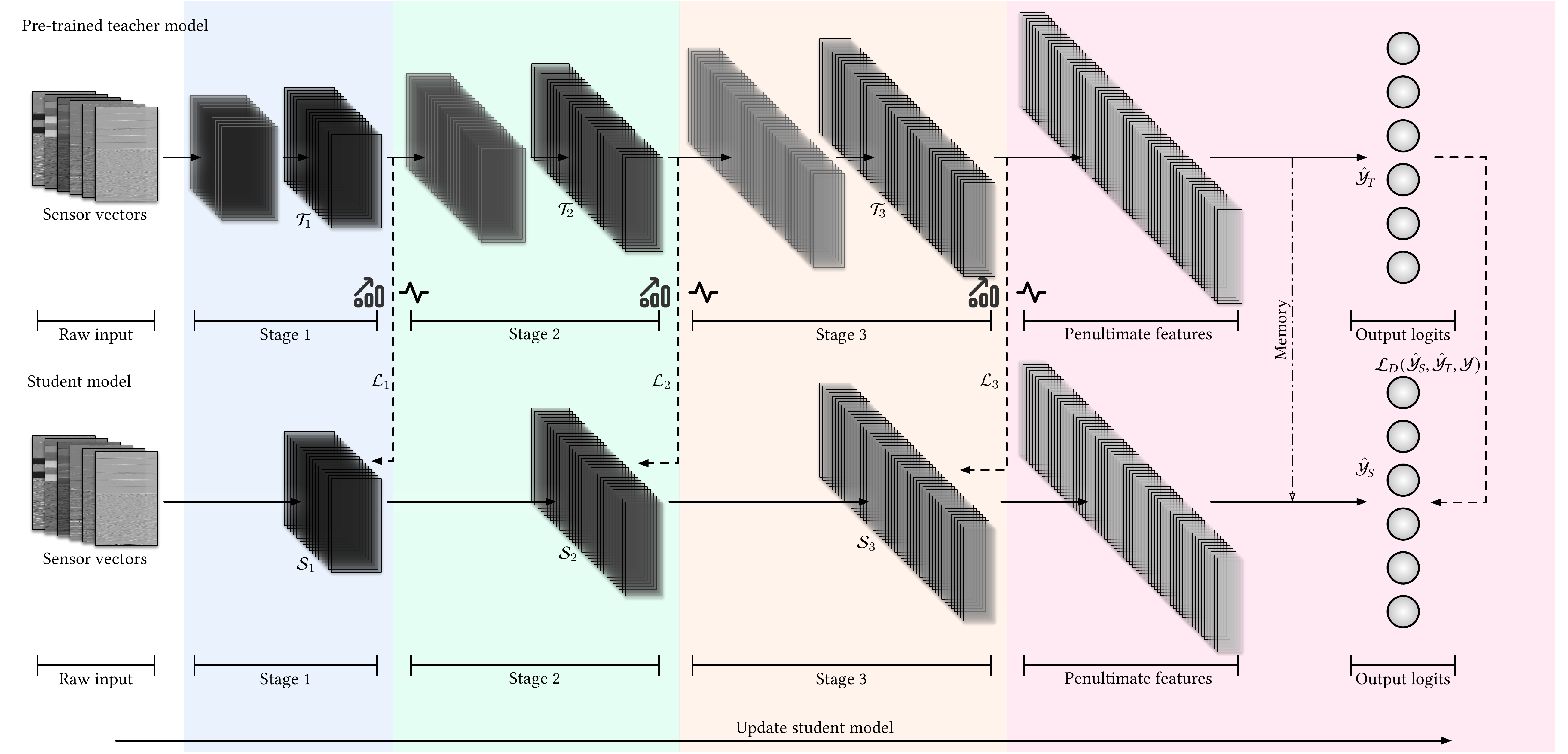}
	\caption{The pipeline of Stage-Memory-Logits Distillation (SMLDist) for HAR.}
	\label{fig:skd-method}
\end{figure*}

This section provides a detailed presentation of the proposed Stage-Memory-Logits Distillation (SMLDist) methods. We begin by introducing the problem definitions and notations. Following the order of Stage-Memory-Logits Distillation, the sections \ref{sec:method}.A to \ref{sec:method}.C will present stage distillation, memory distillation, and logits distillation. Figure \ref{fig:skd-method} illustrates the pipeline of stage distillation and logits distillation for constructing a four-stage model (three stages for representations and one final stage for logits).

SMLDist is a multi-knowledge distillation based method for HAR modeling. The HAR model is represented by a function $f: \mathcal{X}\rightarrow\mathcal{Y}$, which maps the raw sensor signal vectors $\mathcal{X}$ to the final class logits $\mathcal{Y}$. This mapping is a vital component of the entire method. Human activity prediction and classification are performed by applying the HAR method to each temporal window of the sensor signals. For instance, for the $i$-th window in the dataset, the raw signal vectors are denoted as $\mathcal{X}_i$, and the corresponding class logits as $\mathcal{Y}_i$. The collection of all human activity classifications is denoted as $C=\{c_i\}_{i=1}^{n}$. Typically, the softmax function $\sigma(\cdot)$ is used to convert the class logits $\mathcal{Y}$ into the class probability vector $P=(P_{c_1}, \cdots, P_{c_n})$. To achieve a balance between the accuracy and efficiency of the HAR model, we perform stage knowledge distillation on the deep vanilla model and leverage the self-adaptive intuition-memory model to enhance the model's recognition efficiency.

\subsection{Stage Knowledge for HAR}

Stage knowledge refers to the feature knowledge provided by the intermediate layers of the teacher model, which includes the environmental context, motion trends, and motion posture learned by the teacher model from the raw data. By effectively utilizing feature distillation methods specifically designed for HAR, pre-training of filters in the student model can be performed, resulting in improved filter quality and enabling the model to learn knowledge extraction capabilities similar to that of the teacher model.

By analyzing the teacher and student models, we can identify the periodic characteristics (such as motion posture) and tendency characteristics (such as environmental context and motion trends) present in HAR samples. Figure \ref{fig:cam} illustrates the comparison of CAM (Class Activation Map) \cite{DBLP:conf/cvpr/ZhouKLOT16}, which highlights the salient points in the model's stage 1 features. The three sensor samples ("walking downstairs," "sitting," and "sit to lie" from the HAPT \cite{hapt} dataset) exhibit distinct periodicity and tendencies. The "walking downstairs" activity demonstrates a typical periodic nature with a specific movement pattern. The teacher model emphasizes localized high-frequency features. Similarly, the "Sitting" and "Sit to lie" activities display inherent frequency relationships in their features. Human activities exhibit unique frequency characteristics. The teacher model captures periodic features associated with user behavior modes. By implementing the interaction between the knowledge provided by the teacher and the student from a frequency-domain perspective, it is possible to effectively facilitate the student's learning of these periodic patterns. Therefore, the student model needs to further explore the periodicity of the samples.

\begin{figure}[]
    \begin{algorithm}[H]
        \caption{Stage distillation}
        \label{alg:stage-distillation}
        \begin{algorithmic}[1]
            \Require Dataset $X$, teacher $f_T={f_T}_1\circ {f_T}_2\circ\cdots\circ {f_T}_n\circ \mathcal{H}_T$, distilling optimizer $O_i$ and loss function $L_i$ in stage $i$.
            \Ensure Student $f_S={f_S}_1\circ {f_S}_2\circ\cdots\circ {f_S}_n\circ \mathcal{H}_S$.
            \State Initialize model $f_S$.
            \For{stage $i=1,2,\cdots,n$} \Comment Stage $i$.
            \State Let $F_T={f_T}_1\circ\cdots\circ{f_T}_i$ and $F_S={f_S}_1\circ\cdots\circ{f_S}_i$.
            \For{each epoch until $F_S$ performs well}
            \For{each batch $\mathcal{X}$ \textbf{in} $X$}
            \State $\mathcal{T}_i=F_T(\mathcal{X})$
            \State $\mathcal{S}_i=F_S(\mathcal{X})$
            \State Back-propagate with $L_i(\mathcal{S}_i,\mathcal{T}_i)$.
            \State Optimize $F_S$ with $O_i(\nabla F_S)$.
            \EndFor
            \EndFor
            \EndFor
        \end{algorithmic}
    \end{algorithm}
\end{figure}

Let's assume a model $f=f_1\circ \cdots\circ f_n\circ h$ consisting of $n$ stages. In this model, the $i$-th stage $f_i$ extracts hidden features with different scales of perceptive fields, while the final layer $h$ classifies activities based on the features it extracts from the previous stages. As we increase $i$, stage $f_i$ gains the ability to perceive features on a larger temporal scale. The classifier utilizes the features perceived from previous stages for the final classification. To guide the student model, we use the features $\mathcal{T}_i=({f_T}_1\circ \cdots\circ {f_T}_i)(\mathcal{X})$ produced by the teacher model's $i$-th stage. These features instruct the corresponding student stage to extract features $\mathcal{S}_i=({f_S}_1\circ \cdots\circ {f_S}_i)(\mathcal{X})$. In stage distillation, we strictly constrain the corresponding student features $\mathcal{S}_i\in \mathbb{R}^{C_i\times L_i}$ to match the teacher features $\mathcal{T}_i\in \mathbb{R}^{C_i\times L_i}$ by guiding the student stages $({f_S}_1\circ \cdots\circ {f_S}_i)(\cdot)$ to mimic the mapping $\mathcal{X} \rightarrow \mathcal{T}_i$ using the loss function $L_i$.

We define the loss function $\mathcal{L}_i$ as follows:
\begin{equation}
    \mathcal{L}_i=\underbrace{\frac{1}{C_i\times L_i}\left\Vert\mathrm{rfft}(\mathcal{T}_i)-\mathrm{rfft}(\mathcal{S}_i)\right\Vert_2}_\text{Periodic term} + \underbrace{\frac{1}{C_i\times L_i}\left\Vert\mathcal{T}_i-\mathcal{S}_i\right\Vert_2}_\text{Tendency term},
\end{equation}
where $\mathrm{rfft}(\cdot)$ represents the one-dimensional Fourier transform of real-valued input.

As shown in Algorithm \ref{alg:stage-distillation}, the student starts the learning process for the next stage at the conclusion of each stage.

The teacher model captures the temporal characteristics of human activity through pre-training. Human activity exhibits periodicity and tendency in the representation of sensor sequences. Due to the structural characteristics of convolutional networks, these temporal characteristics can be preserved in the intermediate stages. Consequently, we can analyze the intermediate-stage features of the convolutional network in both the time domain and the frequency domain. By distilling the knowledge in the frequency domain, the student model can capture gait switching, an intuitive representation of walking activity. Additionally, the student model can focus on the low-frequency relationship apparent in the samples of activities "Sitting" and "Sit to lie" shown in Figure \ref{fig:cam}. The tendency term in $L_i$ indicates the similarity of the distributions between $\mathcal{T}_i$ and $\mathcal{S}_i$ in the temporal sequence, representing the similarity in dynamic tendencies. Periodic knowledge assists the students in understanding data tendencies from a periodic perspective. Therefore, we introduce an additional term to distill the frequency domain knowledge from the intermediate-stage features.

SMLDist employs a progressive training process. At each stage, student models aim to achieve maximum consistency with their teacher. The use of staged learning objectives prevents student models from becoming distracted by localized features. We consider the final learning stage as the classifier training phase. This stage determines the model's ability to connect high-dimensional features with class distribution, which is crucial for the classification task. To enhance the final stage of learning, we combine memory distillation with logits distillation.

\subsection{Memory Knowledge \& Auto-search of Heterogeneous Heads for Teachers and Students }

\begin{figure}
	\centering
	\includegraphics[width=0.45\textwidth]{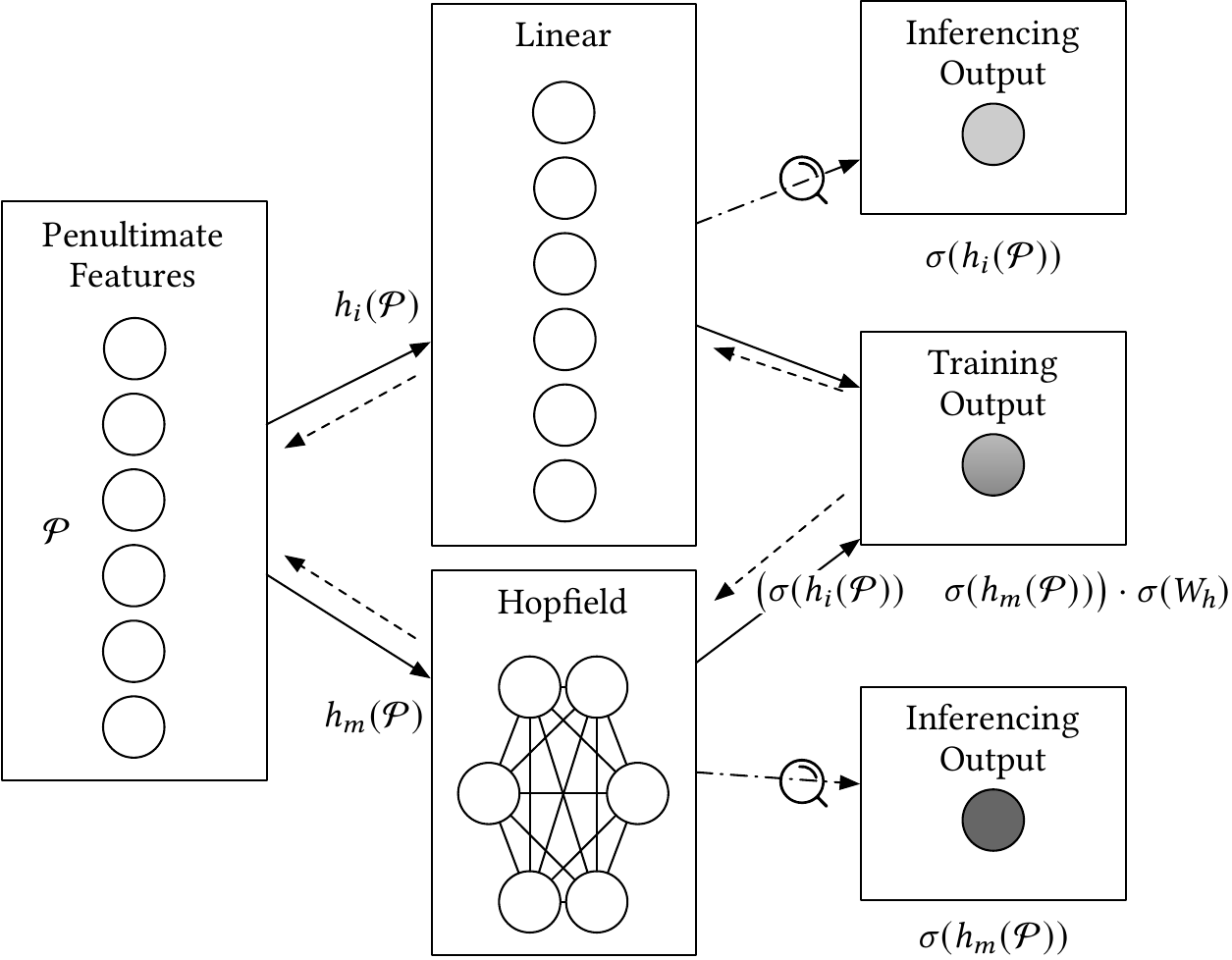}
	\caption{Competitive training and automatic search for classifiers. Within the SMLDist pipeline, the gradient adjusts the importance of each classifier, allowing us to select the classifier with the highest importance for deployment in the final model.}
	\label{fig:intuition-memory}
\end{figure}

The classifier in a deep neural network plays a crucial role in specific tasks. The overall memory of the classifier comprises its structure and parameters. In the context of HAR, we investigate the knowledge contained within the classifier's structure and parameters. One limitation of feature-based distillation is that student models lack the ability to make decisions based on mimicking the teachers' features. To address this issue, we introduce multiple heterogeneous classifiers to the model, enabling them to compete during the training process. We incorporate a modern Hopfield-based classifier \cite{ramsauer2020hopfield} to compete against the plain linear classifier, introducing heterogeneity among different classifiers as depicted in Figure \ref{fig:intuition-memory}. Theoretically, for each classifier $h$ in the set $\mathcal{H}$, the probability of it being the best choice is denoted as $q_h$. To approximate this probability, we assign a learnable weight $\mathcal{W}_h\in\mathcal{W}$ to each $h$, and they satisfy $\hat{q}_h = \sigma(\mathcal{W}_h|\mathcal{W})$, where $\sigma$ denotes the softmax function. During the training process, we utilize the composed response $\hat{\mathcal{Y}}$ from all $h\in\mathcal{H}$, given by
\begin{equation}
    \hat{\mathcal{Y}}=\sum_{h\in\mathcal{H}}h(\mathcal{P})\hat{q}_h,
\end{equation}
where $\mathcal{P}=(f_{1}\circ\cdots\circ f_{n})(\mathcal{X})$. In Figure \ref{fig:intuition-memory}, $\hat{q}_h$ is represented as $\hat{q}_h\sigma(W_h)$. As the optimization process progresses, the classifier with better expressive ability for the specific HAR task gradually gains a higher probability $\hat{q}_h$. We consider the probability $\hat{q}_h$ to reflect the importance of the classifier.

In the final stage, the student no longer mimics the teacher's extracted features but directly inherits the teacher's memory. By inheriting this memory, we can fine-tune the student to align the classifier with the overall model, typically requiring only a few epochs. Following the fine-tuning process, we identify the classifier with the highest estimated importance and remove any redundant classifiers. Additionally, we distill the knowledge of logits into the student during the fine-tuning process. The inherited memory and soft targets serve as a bridge between the learned representation and unexplored class distribution, as illustrated in Algorithm \ref{alg:memory-logits-distillation}.

\begin{figure}[]
    \begin{algorithm}[H]
        \caption{Memory \& logits distillation of the final stage}
        \label{alg:memory-logits-distillation}
        \begin{algorithmic}[1]
            \Require Dataset $X$ with labels $Y$, teacher $f_T={f_T}_1\circ {f_T}_2\circ\cdots\circ {f_T}_n\circ \mathcal{H}_T$, student $f_S={f_S}_1\circ {f_S}_2\circ\cdots\circ {f_S}_n\circ \mathcal{H}_S$ pre-trained by stage distillation, distilling optimizer $O_i$ and loss function $L=\mathcal{L}_{D}$.
            \Ensure Student $f_S$.
            \State Clone memory from $\mathcal{H}_T$ to $\mathcal{H}_S$. \Comment Memory distillation
            \For{each epoch until $F_S$ performs well}
            \For{each batch $\mathcal{X},\mathcal{Y}$ \textbf{in} $X,Y$}
            \State $\hat{\mathcal{Y}}_T=f_T(\mathcal{X})$
            \State $\hat{\mathcal{Y}}_S=f_S(\mathcal{X})$
            \State Back-propagate with $L(\hat{\mathcal{Y}}_S,\hat{\mathcal{Y}}_T,\mathcal{Y})$.\Comment Logits distillation
            \State Optimize $F_S$ with $O_i(\nabla f_S)$.
            \EndFor
            \EndFor
            \State $h={\mathcal{H}_S}_{\mathrm{argmax}(\hat{q}_{\mathcal{H}_S})}$. \Comment Auto-search.
            \State Replace $\mathcal{H}_S$ with $h$ in $f_S$.
        \end{algorithmic}
    \end{algorithm}
\end{figure}

\subsection{Logits Knowledge}

Vanilla knowledge distillation \cite{hinton2015distilling} transfers the generalization ability from a cumbersome model to a smaller model by distilling the knowledge contained in the teacher model's logits predictions to the smaller model. Teachers provide students with more domain-specific knowledge, enabling them to solve practical problems. The soft target, which incorporates hidden relations between classes, contains more information than raw one-hot labels. It captures the non-independence of classes and uncovers hidden similarities explained by the teacher model. In vanilla class probability distillation, we denote the output class logits as $\hat{\mathcal{Y}_T}$, with corresponding probabilities $P(\hat{\mathcal{Y}_T})_c$. For a class logits $\hat{\mathcal{Y}}$, we calculate the probability $P(\hat{\mathcal{Y}})_c$ using softmax function $\sigma$. The vanilla class probability distillation employs both the soft target from the cumbersome teacher model and the manually labeled hard target. When distilling the soft target to the student model, the training loss $\mathcal{L}_{KD}$ combines the cross-entropy between the predicted logits $\hat{\mathcal{Y}}_S$ and the soft target $\hat{\mathcal{Y}}_T$, as well as the cross-entropy between the predicted logits $\hat{\mathcal{Y}}_S$ and the hard ground-truth one-hot label $\mathcal{Y}$. The combined loss $\mathcal{L}_{H}$ of the vanilla class probability distillation is shown in Equation \ref{equ:hinton-cross-entropy}:
\begin{equation}
	\label{equ:hinton-cross-entropy}
	\mathcal{L}_{H}(\hat{\mathcal{Y}}_S,\hat{\mathcal{Y}}_T,\mathcal{Y})=\mathcal{L}_{CE}(\hat{\mathcal{Y}}_S, \mathcal{Y}_i) + \lambda\mathcal{L}_{CE}(\frac{\hat{\mathcal{Y}}_S}{\tau},\frac{\hat{\mathcal{Y}}_T}{\tau})\\
\end{equation}
where $\mathcal{L}_{CE}$ is the cross-entropy loss function, the temperature $\tau$ controls the relaxation ratio, and the parameter $\lambda$ is the weight that balances the influence of the hard label and the soft target.

Logits distillation ensures the semantic learning goals of the student model. The hard one-hot labels are not always ground truth in the real situation \cite{feature-normalized-kd}. Manually labeled hard labels may introduce new noise for the HAR tasks. The movement features of the user can include many patterns of activity. For example, when running upstairs, the user's activity state should be a combination of several simple activities, including jumping, running, or sometimes walking. The one-hot method to express the actual activity possibility value may ignore some fundamental activities. When the state is labeled as "going upstairs", other related activities such as "walking" or "running" should also gain certain possibilities. However, one-hot labeled logits cannot express those relatively secondary activity classes. Therefore, a harder label may be more likely to train over-fitted activity recognizing models \cite{label-smooth-help?}.

The soft targets provided by the teacher model reveal the implicit inter-class correlations. However, the teacher model cannot always provide accurate predictions for possible human activity classes. To address this, we employ a conditional control technique, denoted as $Q(\cdot)$, which leverages the manually labeled class to guide the student model \cite{conditional-teacher-student-learning}. The $c$-th element of the controlled probability provided by the teacher model is denoted as $Q(\hat{\mathcal{Y}}_T)_c$,
\begin{equation}
	\begin{array}{c}
		R(\hat{\mathcal{Y}}_T)_c=\left\{
		\begin{array}{ll}
			\gamma,                   & \mathrm{argmax}\ \mathcal{Y}\ \neq \mathrm{argmax} \ \hat{\mathcal{Y}}_T \\
			P(\hat{\mathcal{Y}}_T)_c, & \text{otherwise}
		\end{array}
		\right. ,\\
		Q(\hat{\mathcal{Y}}_T)=\sigma(R(\hat{\mathcal{Y}}_T)),
	\end{array}
\end{equation}
where $\gamma$ represents the hardness factor of $R(\cdot)$. The controlled probability $Q(\hat{\mathcal{Y}}_T)$ corrects the effects of mislabeled classes and smooths the probability distribution of soft targets. Subsequently, we use $\mathcal{L}_D(\cdot)$ as the loss function to train the front-end classifier of the student model after training the previous stages.
\begin{equation}
	\begin{split}
		\mathcal{L}_{D}(\hat{\mathcal{Y}}_S,\hat{\mathcal{Y}}_T,\mathcal{Y})=&-(\sum_{i=1}^{C}\mathcal{Y}_i\log(P(\hat{\mathcal{Y}}_S)_i)\\
		&+\lambda\sum_{i=1}^{C}Q(\frac{\hat{\mathcal{Y}}_T}{\tau})_i\log(P(\frac{\hat{\mathcal{Y}}_S}{\tau})_i)).\\
	\end{split}
\end{equation}

SMLDist is a robust knowledge distillation pipeline that comprises stage distillation, memory distillation, and logits distillation. It aims to enhance the HAR performance of plain convolutional deep models. By leveraging multi-knowledge distillation, SMLDist achieves significant improvements in the HAR performance of lightweight models.

\section{Experimental Evaluation}

\label{section:experimental-evaluation}

In this section, we comprehensively benchmark and analyze the performance of SMLDist through extensive experiments conducted on various public HAR datasets. Our benchmarking results showcase the effectiveness of SMLDist and underscore the significance of multi-knowledge distillation in the context of HAR.

\subsection{Prerequisites}

\begin{table*}[t]
    \caption{Description of public HAR datasets in various challenging scenarios. ("U": generalization of users. "S": generalization of sensor displacements. "D": generalization of diverse activity modes.)}
    \label{tab:datasets-overall}
    \centering
    \begin{tabular}{c|cccccc|c}
        \toprule
        Name          & Subjects & Activities & Body positions & Sensors & Sample Rate & Window size & Scenarios\\
        \midrule
        RealWorld-HAR & 15       & 8          & 7              & 3       & 45Hz        & 5 seconds & U   \\
        UCI-HAR       & 30       & 6          & 2              & 2       & 50Hz        & 2.56 seconds & U \\
        DSADS         & 8        & 19         & 5              & 3       & 50Hz        & 5 seconds  & U   \\
        HAPT          & 30       & 12         & 1              & 2       & 50Hz        & 3 seconds  & U   \\
        REALDISP      & 17       & 33         & 9              & 3       & 50Hz        & 3 seconds & S   \\
        OPPOTUNITY & 4 & 17 & 5 & 3 & 30Hz & 2 seconds &D \\
        HTC-TMD       & 224      & 10         & 3              & 3       & 47Hz        & 5 seconds & D   \\
        Skoda & 1 & 10 & 10 & 1 & 98Hz & 2 seconds & D \\
        \bottomrule
    \end{tabular}
\end{table*}

We conduct experiments to evaluate the performance of our method compared to baseline methods in multiple aspects, such as accuracy and efficiency, for HAR metrics. As displayed in Figure \ref{tab:datasets-overall}, we use diverse datasets to confirm the effectiveness of our method in various challenging scenarios, including the generalization of users, the generalization of sensor displacements, and the recognition of diverse activity modes. We conduct comprehensive testing on representative datasets with abundant samples, covering metrics including accuracy and efficiency, including RealWorld-HAR, UCI-HAR, HTC-TMD, HAPT, DSADS, and REALDISP. For other datasets, we only evaluate HAR methods for accuracy. For comprehensive information on these datasets, refer to Table \ref{tab:datasets-overall}. For details concerning the server and embedded environments used in our experiments, refer to Table \ref{tab:evaluation-environments}. The general hyperparameter configurations that we use during the training process are detailed in Table \ref{tab:hyperparams}.

\textbf{Generalization of users.} Activity patterns among different users exhibit significant variations. The ability of a HAR model to estimate HAR performance on users outside the training set is a critical indicator of the model's effectiveness. We select several HAR datasets for our tests of accuracy and efficiency, where the test set comprises users not covered in the training set, including RealWorld-HAR \cite{realworld-har}, HAPT \cite{hapt}, DSADS \cite{DSADS-1,DSADS-2,DSADS-3}, and UCI-HAR \cite{uci-har}. The users in the test set show substantial differences from those in the training set in terms of age, height, weight, and habits, which adequately reflects the data distribution variance from the training set. This distribution variance is pivotal in demonstrating whether the model can accurately categorize common activity patterns in HAR, serving as a representative validation of the model's effectiveness in HAR.

\textbf{Generalization of sensor displacements.} The displacement of wearable sensors significantly impacts the sensor measurements. In real-world scenarios, users wearing sensors improperly can lead to a deviation in sensor measurements from those obtained under ideal wearing conditions. We utilize the REALDISP \cite{REALDISP-1,REALDISP-2} dataset to validate this challenging scenario, where the training set contains only data from ideal wearing conditions, while the test set is significantly influenced by individual wearing styles. Methods that demonstrate superior performance in this scenario exhibit enhanced generalization capabilities towards variations in sensor displacements.

\textbf{Recognition of diverse activity modes.} We validate various methods under more diverse HAR application scenarios. We select benchmarks with richer activity divisions to verify HAR's broader application in terms of accuracy and efficiency. The HTC-TMD \cite{htc-tm} dataset categorizes common transportation modes, testing the HAR methods' ability to recognize the transportation mode a user is in. We further validated the accuracy of various methods using benchmarks in more specific scenarios. The OPPORTUNITY \cite{OPPOTUNITY-1,OPPOTUNITY-2} dataset categorizes the daily activities of users into finer-grained classes, focusing on more life-like activities. The Skoda \cite{Skoda-1,Skoda-2} dataset provides a detailed division of behaviors of automobile workers in a car factory setting. These benchmarks adequately test the performance of different HAR methods in recognizing challenging, finely categorized activities.


We design our benchmarks for a wide range of IMU datasets. IMU data, influenced by factors such as the data collection environment and devices, exhibits numerous outliers that deviate from the normal numerical range. To ensure a fair and trustworthy comparison of test results, we apply the robust min-max scaling method $S_\mathrm{robust}$ to preprocess the sensor measurements $\mathcal{X}$ as
\begin{equation}
	\label{equ:robust-scaling}
	\begin{split}
		\mathrm{IQR}(\mathcal{X})&= \mathrm{Q_3}(\mathcal{X}) - \mathrm{Q_1}(\mathcal{X}), \\
		L_{\mathrm{lower}}(\mathcal{X})&= \mathrm{Q_1}(\mathcal{X})-1.5\cdot\mathrm{IQR}(\mathcal{X}), \\
		L_{\mathrm{upper}}(\mathcal{X})&= \mathrm{Q_3}(\mathcal{X})+1.5\cdot\mathrm{IQR}(\mathcal{X}), \\
		S_{\text{robust}}(\mathcal{X})&=\frac{\mathrm{clip}(\mathcal{X},L_{\text{lower}}(\mathcal{X}),L_{\text{upper}}(\mathcal{X}))}{4\cdot\mathrm{IQR}(\mathcal{X})}, \\
	\end{split}
\end{equation}
where $\mathrm{Q_1}$ represents the first quartile of the raw sensor values $\mathcal{X}$, $\mathrm{Q_3}$ represents the third quartile of raw sensor values $\mathcal{X}$, and $\mathrm{IQR}$ denotes the interquartile range of raw sensor values $\mathcal{X}$. $S_\mathrm{robust}$ ensures a more evenly distributed numerical range for the input IMU data into the model, enabling us to concentrate our assessment on the HAR patterns.

\begin{table}[]
	\caption{Evaluation environments of SMLDist.}
	\centering
	\begin{tabularx}{0.45\textwidth}{c|cccc}
		\toprule
		Type                                                            & \multicolumn{4}{c}{Information}                                                                     \\
		\midrule
		\multirow{6}{*}{\rotatebox{90}{Server}}                         & \multirow{2}{*}{\rotatebox{90}{CPU}}      & Model                            & Frequency   & Memory \\
		                                                                &                                           & Intel Xeon Gold 6230             & 2.10GHz     & 187GiB \\
		                                                                & \multirow{2}{*}{\rotatebox{90}{GPU}}      & Model                            & Performance & Memory \\
		                                                                &                                           & NVIDIA Tesla V100S               & 130TFLOPS   & 32GiB  \\
		                                                                & \multicolumn{2}{c}{Operating system}      & \multicolumn{2}{c}{Architecture}                        \\
		                                                                & \multicolumn{2}{c}{CentOS Linux 7.4.1708} & \multicolumn{2}{c}{amd64}                               \\
		\midrule
		\multirow{6}{*}{\rotatebox{90}{\begin{tabular}{c}
					Embedded \\
					device   \\
				\end{tabular}}} & \multirow{2}{*}{\rotatebox{90}{CPU}}      & Model                            & Frequency   & Memory \\
		                                                                &                                           & NVIDIA Carmel                    & 2.30GHz     & 32GiB  \\
		                                                                & \multirow{2}{*}{\rotatebox{90}{GPU}}      & Model                            & Performance & Memory \\
		                                                                &                                           & NVIDIA Volta                     & 11TFLOPS    & 32GiB  \\
		                                                                & \multicolumn{2}{c}{Operating system}      & \multicolumn{2}{c}{Architecture}                        \\
		                                                                & \multicolumn{2}{c}{Ubuntu 18.04.5 LTS}    & \multicolumn{2}{c}{aarch64}                             \\
		\bottomrule
	\end{tabularx}
	\label{tab:evaluation-environments}
\end{table}

\begin{table}[t]
    \centering
    \caption{Global hyperparameter configurations for SMDList.}
    \begin{tabular}{c|cccc}
    \toprule
    \makecell{Hyperparameters}&\makecell{Learning\\Rate}&\makecell{Batch\\Size}&\makecell{PyTorch\\Version}\\
    \midrule
    Value&$1\times {10}^{-4}$&256&1.7.1\\
    \bottomrule
    \end{tabular}
    \label{tab:hyperparams}
\end{table}

In our benchmark, we need to perform performance comparisons of HAR models constructed under different model structures and distillation methods from the perspectives of classification accuracy, computational complexity, and energy expenditure. HAR tasks require accurate assessment of multi-class recognition accuracy, necessitating a balanced evaluation of multi-class results. Among the various computation methods for F1 scores, the F1 macro score accurately highlights multi-class recognition accuracy. Therefore, for the classification performance of HAR models, we utilize Accuracy and F1 macro score as the metrics. We conduct a comparison of computational complexity and energy expenditure for selected dataset configurations. The metrics for computational complexity encompass the number of parameters and multiply-accumulates (MACs), corresponding to spatial complexity and temporal complexity, respectively. Using power estimation on the NVIDIA Jetson AGX Xavier, we assess the equivalent daily energy consumption for different model structures and dataset configurations. We analyze the configurations of the SMLDist models involved in the comparison in Section \ref{sec:scale-efficiency}.

\subsection{Evaluation in Challenging Scenarios}

We validate the performance of our method's trained baseline model (MobileNet V3 \cite{mobilenetv3}) in various challenging scenarios, including the generalization of users, generalization of sensor displacement, and recognition of diverse activity modes. Our approach demonstrates the capability to outperform different state-of-the-art model architectures, maintaining strong predictive performance even in challenging scenarios. In our benchmark, we implement EmbraceNet, IndRNN, and Dynamic-WHAR using the original open-source code, while we independently re-implemented Global-Fusion, SparseSense, AttnSense, and DeepConv-LSTM.

\begin{figure}
    \centering
    \includegraphics[width=\columnwidth]{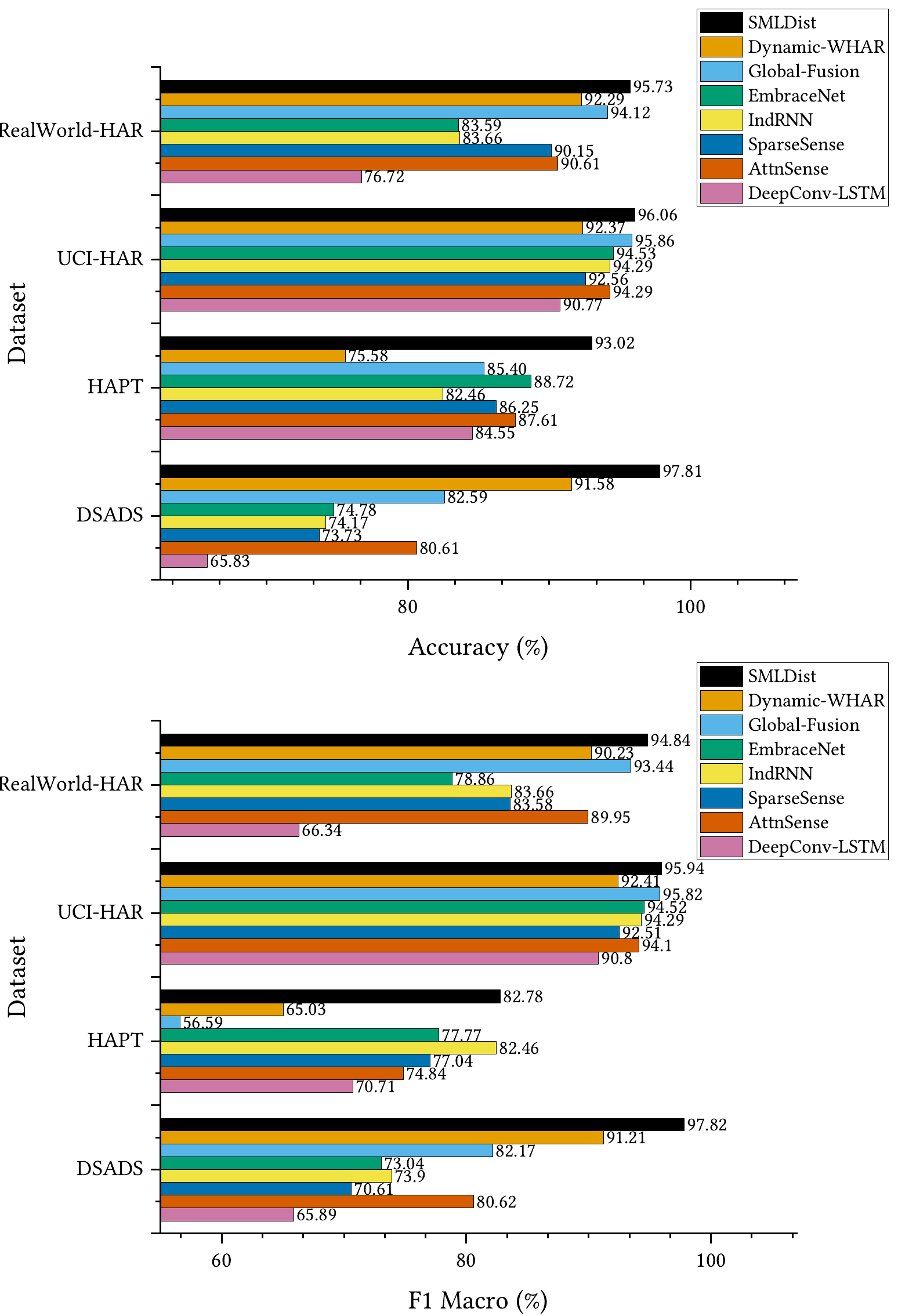}
    \caption{Comparison of Accuracy and F1 macro between SMLDist and other state-of-the-art HAR model architectures, focusing on the generation of users, using the test sets comprising data from users not included in the training set.}
    \label{fig:gou-total}
\end{figure}

An effective HAR method should adapt to diverse user wearables while consistently delivering accurate recognition outcomes. To assess the utility of various model structures and methods, we conduct benchmarks using samples from users not involved in the model training phase, and the results are shown in Figure \ref{fig:gou-total}. In general, for HAR models, the greater the number of introduced sensors and the more common recognition categories in the identified scene, the stronger the model's generalization ability to different users. DSADS contains abundant daily activity categories, while HAPT includes a variety of transition activity categories. In such scenarios, extracting common activity patterns across different users becomes more challenging. Our SMLDist demonstrates the capability to adapt to variations in activities resulting from different user habits and outperforms other state-of-the-art models in generalization performance on unfamiliar user samples. 

\begin{figure}
    \centering
    \includegraphics[width=0.95\columnwidth]{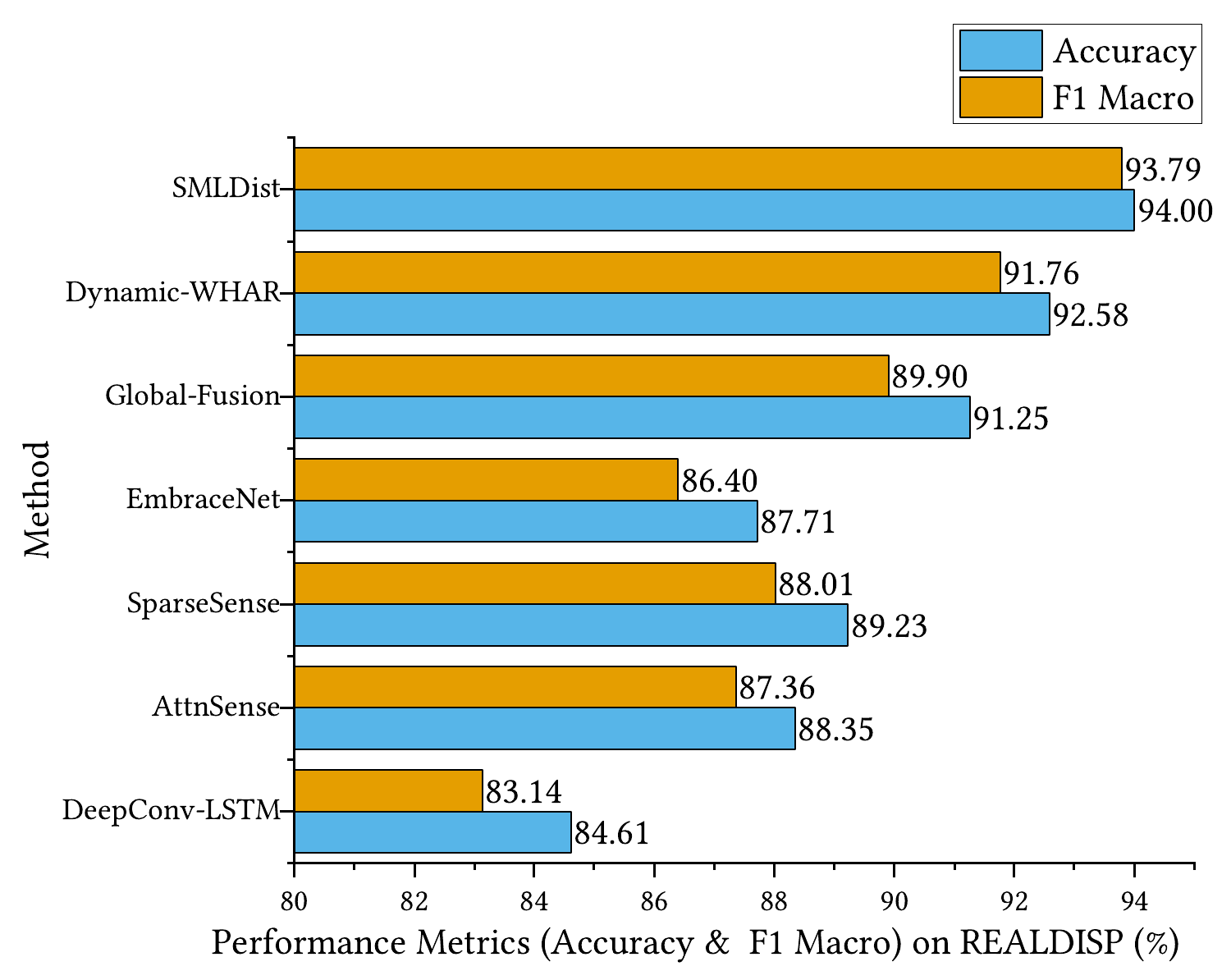}
    \caption{Comparison of Accuracy and F1 macro between SMLDist and other state-of-the-art HAR model architectures, focusing on the generation of sensor displacements, using the test sets comprising self-placement and induced-displacement, which are completely distinct from the ideal-placement in the training set.}
    \label{fig:god-total}
\end{figure}

Unlike differences in activity patterns among users, variations in the way the same user wears sensors can also pose challenges for HAR recognition. REALDISP introduces an evaluation of HAR model recognition performance under different wear configurations, providing a clear reflection of this scenario. Using REALDISP, we conduct the performance evaluation for various model architectures in the context of the generalization of sensor displacement scenarios. REALDISP benchmarking reflects the adaptability of different model structures to various user wear configurations. As seen in Figure \ref{fig:god-total}, our SMLDist significantly assists conventional models in achieving state-of-the-art performance under different wear configurations. 

In addition to recognizing common activities, HAR models need to accurately identify more specific activity categories to realize their potential value in future downstream applications. We compare models trained based on SMLDist with some excellent HAR model architectures in various downstream tasks. These datasets encompass classification scenarios such as fine-grained household modes, transportation modes, and factory work modes. In these scenarios, user behavior patterns are more complex, environmental factors are more intricate, and the application value is higher. Achieving accurate recognition in these scenarios effectively demonstrates the ability of HAR methods to extract knowledge related to human activity patterns and more complex environmental influences.

\begin{figure}
    \centering
    \includegraphics[width=\columnwidth]{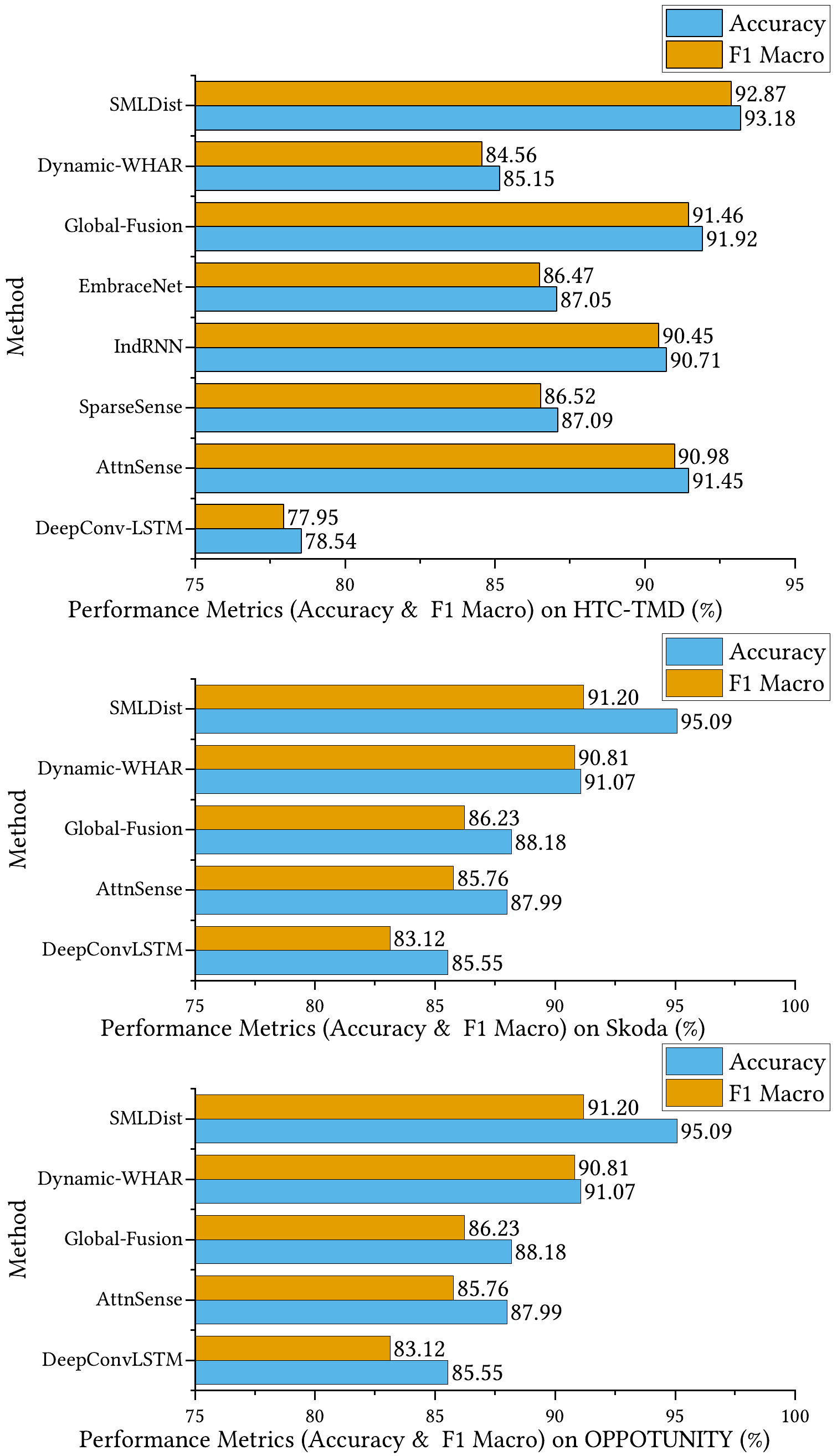}
    \caption{Comparison of Accuracy and F1 macro between SMLDist and other state-of-the-art HAR model architectures, focusing on the recognition of diverse activity modes. This comparison uses test sets designed to evaluate the model's ability to accurately identify a wide range of human activities.}
    \label{fig:div-har-total}
\end{figure}

\begin{table}[t]
    \centering
    \caption{Most \& least successful activities recognized by SMLDist.}
    \begin{tabular}{c|c|c}
    \toprule
    Datasets & Top 3 modes & Bottom 3 modes \\
    \midrule
    \multirow{3}{*}{REALDISP} & Rowing & Knees to the breast \\
    & Rotation on the knees & Frontal hand claps \\
    & Waist bends forward & Jump up \\
    \midrule
    \multirow{3}{*}{DSADS} & Sitting & Lying on right side \\
    & Rowing & Lying on back \\
    & Exercising on a stepper & Standing in an elevator still \\
    \midrule
    \multirow{3}{*}{OPPOTUNITY} & Drink Cup & Close Drawer3 \\
    & Open Fridge & Open Drawer 1 \\
    & Toggle Switch & Close Drawer 1 \\
    \bottomrule
    \end{tabular}
    \label{tab:top-bottom-modes}
\end{table}

Figure \ref{fig:div-har-total} compares state-of-the-art HAR models in scenarios like transportation modes, factory work activities, and daily household activities, using HTC-TMD, Skoda, and OPPOTUNITY datasets. We reference performance metrics from baseline model architectures in \cite{10.1145/3550331} and employ the same dataset splits to train SMLDist models. SMLDist showcases superior classification accuracy in these scenarios, highlighting its potential for precise recognition in HAR tasks and emphasizing its significant practical value.

We direct our attention to the two datasets with the highest number of activity modes detailed in Table \ref{tab:datasets-overall}, specifically REALDISP, DSADS, and OPPOTUNITY. We continually rank the accuracy of SMLDist for each activity mode, distinguishing the top 3 and bottom 3 activity modes, as illustrated in Table \ref{tab:top-bottom-modes}. In our SMLDist model, activities with more distinct periodic patterns are accurately identifiable, such as "rowing" or "rotation on the knees" in the REALDISP dataset. In the OPPOTUNITY dataset, the top three most accurately recognized activity patterns exhibit minimal similarity to other activities, thereby reducing confusion. Conversely, the poorest performing activity patterns often share similarities with multiple categories, leading to misclassification, as exemplified by "close drawer 1" and "close drawer 3". Additionally, environmental motion exerts a significant influence, such as the "standing in an elevator still" scenario in the DSADS dataset.

Achieving state-of-the-art performance across scenarios such as generalization of users, generalization of sensor displacements, and recognition of diverse activity modes underscores the effectiveness and versatility of our training methodology. This consistent success demonstrates our approach's ability to handle a wide range of human activities, sensor variations, and user differences, proving its superiority over conventional model architectures. Such outcomes suggest that our training method not only enhances model adaptability to unseen data but also significantly improves recognition accuracy in diverse real-world situations, marking a substantial advancement in the field of HAR.

\subsection{Distillation Methodology: Benchmarks and Ablations}

\begin{table}[tbp]
    \centering
    \setlength\tabcolsep{3pt}
    \caption{Configuration of the teacher models.}
    \label{tab:teacher-models}
    \begin{tabular}{c|ccccc}
    \toprule
    \multirow{2}{*}{Dataset} & MACs      & Parameters & Energy cost\\
    & (M)       & (M)        & ($W\cdot h$/day) \\
    \midrule
    RealWorld-HAR&309.8368&78.2032&1.6711\\
    UCI-HAR&266.2002&82.6504&2.7705\\
    HTC-TMD&308.4033&78.1904&1.6416\\
    HAPT&190.3309&12.8585&0.9219\\
    DSADS&443.6945&25.4578&0.8240\\
    REALDISP&666.6744&72.6468&2.6580\\
    \bottomrule
    \end{tabular}
\end{table}

As an effective deep learning pipeline for HAR tasks based on knowledge distillation, we conduct 
comprehensive benchmarks and ablation experiments focusing on the knowledge distillation methods within SMLDist. We construct teacher models on selected representative datasets, with configurations including computational complexity, spatial complexity, and energy consumption, as shown in Table \ref{tab:teacher-models}.

\begin{table*}[ht]
	\caption{Predicting performance comparison of various KD methods and ablation evaluation of SMLDist with equal compression ratio on public datasets.}
	\label{tab:kd-performance-comparison}
	\centering
	\begin{tabular}{c|cccccc|cccccc}
		\toprule
		\multirow{2}{*}{Method}                        & \multicolumn{6}{c|}{Accuracy (\%)} & \multicolumn{6}{c}{F1 Macro (\%)}                                                                                                                                                                                                                                                                                                                       \\
        & \rotatebox{75}{RealWorld-HAR}      & \rotatebox{75}{UCI-HAR}           & \rotatebox{75}{HTC-TMD} & \rotatebox{75}{HAPT} & \rotatebox{75}{DSADS} & \rotatebox{75}{REALDISP} & \rotatebox{75}{RealWorld-HAR} & \rotatebox{75}{UCI-HAR} & \rotatebox{75}{HTC-TMD} & \rotatebox{75}{HAPT} & \rotatebox{75}{DSADS} & \rotatebox{75}{REALDISP} \\
		\midrule
		Raw Student                                            & 82.52                              & 94.64                             & 92.73                   & 84.15                & 95.53                 & 93.00                    & 82.27                         & 94.58                   & 92.36                   & 42.13                & 95.51                 & 92.86                    \\
		\midrule
		Vanilla KD \cite{hinton2015distilling}          & 90.31                              & 94.70                             & 92.95                   & 85.15                & 93.86                 & 91.69                    & 86.09                         & 94.67                   & 92.70                   & 43.92                & 93.27                 & 93.12                    \\
		CKD \cite{conditional-teacher-student-learning} & 93.21                              & 95.08                             & 92.41                   & 91.12                & 96.05                 & 93.48                    & 89.29                         & 95.03                   & 92.08                   & 71.49                & 95.84                 & 93.22                    \\
		FitNets \cite{fitnets}                          & 90.08                              & 95.52                             & 88.53                   & 86.98                & 95.26                 & 91.85                    & 84.14                         & 95.46                   & 81.50                   & 43.75                & 95.17                 & 91.37                    \\
		NST \cite{DBLP:journals/corr/HuangW17a}         & 92.37                              & 94.94                             & 86.13                   & 84.38                & 95.88                 & 92.83                    & 89.10                         & 94.89                   & 79.37                   & 44.32                & 95.76                 & 91.95                    \\
		FNKD \cite{feature-normalized-kd}               & 86.56                              & 94.94                             & 92.11                   & 87.90                & 96.93                 & 92.84                    & 80.86                         & 94.83                   & 91.72                   & 56.27                & 96.86                 & 92.56                    \\
		SPKD \cite{similarity-preserving}              & 86.18                              & 95.15                             & 92.40                   & 86.14                & 94.65                 & 93.20                    & 69.78                         & 95.11                   & 92.02                   & 45.05                & 94.63                 & 92.67                    \\
		FT \cite{DBLP:conf/nips/KimPK18}                & 84.27                              & 95.76                             & 92.15                   & 86.75                & 95.00                 & 92.46                    & 68.87                         & 95.70                   & 91.79                   & 44.92                & 94.91                 & 91.86                    \\
		\midrule
        SMLDist w/o S & 90.31 & 94.33 & 87.86 & 89.96 & 89.08 & 92.37 & 83.88 & 94.24 & 77.17 & 77.53 & 88.66 & 91.29 \\
        SMLDist w/o M & 94.20 & 95.86 & 91.92 & 91.79 & 96.75 & 93.94 & 93.50 & 95.77 & 80.15 & 80.31 & 96.67 & 93.54\\
        SMLDist w/o L & 94.96 & 95.72 & 92.92 & 91.85 & 96.93 & 93.61 & 94.37 & 95.58 & 93.53 & 80.41 & 96.92 & 93.32 \\
        SMLDist w/o S, M & 87.02 & 94.71 & 91.39 & 89.50 & 88.60 & 93.67 & 74.06 & 94.65 & 90.87 & 74.93 & 87.95 & 93.51\\
        SMLDist w/o S, L & 91.83 & 94.84 & 92.62 & 86.96 & 88.42 & 92.73 & 87.57 & 94.72 & 91.34 & 73.67 & 87.23 & 92.26\\
        SMLDist w/o M, L & 95.50 & 95.92 & 90.26 & 91.98 & 97.25 & 93.64 & 94.81 & 95.81 & 81.04 & 79.87 & 97.38 & 93.43\\
        \textbf{SMLDist}                               & \textbf{95.73}                              & \textbf{96.06}                             & \textbf{93.18}                   & \textbf{93.02}                & \textbf{97.81}                 & \textbf{94.00}                    & \textbf{94.84}                         & \textbf{95.94}                   & \textbf{92.87}                   & \textbf{82.78}                & \textbf{97.82}                 & \textbf{93.79}                    \\
		\bottomrule
	\end{tabular}
\end{table*}

Using these teacher models, we employ SMLDist to build our student models and compare them with student models of the same architecture (MobileNet V3) trained using a range of state-of-the-art knowledge distillation methods under equivalent conditions. We also conduct ablation tests on SMLDist, as presented in Table \ref{tab:kd-performance-comparison}. To ensure fair performance comparison, we ensure that all teacher and student models in different knowledge distillation pipelines have consistent environmental configurations in terms of structure, compression ratio, and dataset. We independently train the raw student model without any knowledge distillation methods. Our experiments demonstrate that SMLDist exhibits better robustness when compressing deep models, leading to less loss in performance compared to other pipelines. We train the above baseline KD pipelines for 100 epochs, while SMLDist achieves comparable performance in just 5 epochs (as shown in Table \ref{tab:kd-performance-comparison}). Both the model representation and distillation efficiency exhibit reasonable improvements in the case of SMLDist.

All components of SMLDist play a critical role. To evaluate the importance of each component, we conduct ablation experiments on SMLDist with identical configurations for both student and teacher models. As shown in Table \ref{tab:kd-performance-comparison}, we perform experiments by eliminating different combinations of stage distillation, memory distillation, and logits distillation. The most significant performance drop occurs when stage distillation is removed, indicating its vital contribution to the overall effectiveness of SMLDist. Stage distillation significantly improves the performance of the raw student models, demonstrating its stability. However, independent stage distillation performs worse than the raw student models and SMLDist. The introduction of memory distillation and logits distillation enhances the stability of independent stage distillation. Furthermore, we find that directly transferring parameters from the teacher to the student, such as in memory distillation, is not ideal. Memory distillation serves as a complementary component to stage distillation. We perform ablation evaluations for all components of SMLDist using a fine-tuning process of 5 epochs. By reusing the classifiers, the fine-tuning process accelerates the student models' learning without deviating from the learning trajectory. However, excessive fine-tuning negatively impacts the model's performance after reusing the teachers' classifiers. Additionally, logits distillation provides a modest performance improvement compared to stage-memory distillation, as it captures more implicit information than the hard labels. The ablation evaluations clearly demonstrate that the cooperative integration of stage distillation, memory distillation, and logits distillation leads to improved generality and accuracy of the lightweight model.

\subsection{Analyzing Model Efficiency Across Multiple Dimensions}
\label{sec:scale-efficiency}

\begin{figure*}[htbp]
    \centering
    \includegraphics[width=0.7\textwidth]{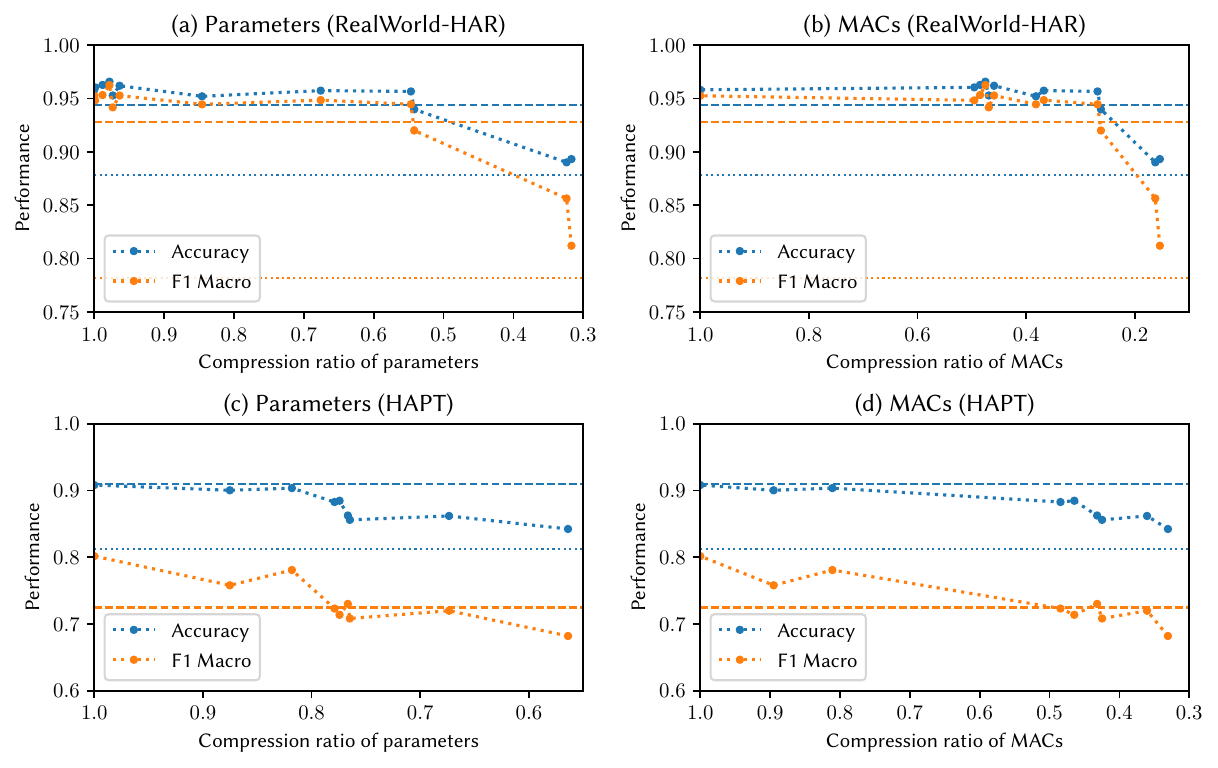}
    \caption{Performance curve of student models with decreasing compression ratio. The dashed line represents the performance metrics achievable by the teacher model without using knowledge distillation, while the dotted line represents the performance metrics achievable by the model configuration with the lowest compression ratio without knowledge distillation.}
    \label{fig:skd-scaling}
\end{figure*}

After verifying the effectiveness of SMLDist, we explore the range within which model compression achieves the best performance-to-cost ratio. We conduct a detailed investigation of different compression ratios for student models based on the RealWorld-HAR and HAPT datasets. As shown in Figure \ref{fig:skd-scaling}, within an appropriate compression ratio range, SMLDist maintains performance metrics similar to the original teacher model. Since the pre-trained teacher model has already extracted valuable knowledge from the original data, it can even slightly improve the performance metrics of similarly sized student models in some cases, surpassing the original teacher model. For models configured with the minimum compression ratio, the performance improvement brought by SMLDist is more significant. We base all the SMLDist models participating in the comparisons in this chapter on the strategy mentioned above, resulting in model configurations that achieve an optimal balance between performance and compression ratio. Thus, we observe that SMLDist serves as a training strategy to enhance performance metrics for the same model configuration, as well as a low-loss model compression strategy.

Based on the selected model configurations, we conducted comparisons of computational complexity, spatial complexity, and energy consumption for different model architectures. We quantified the computational complexity of various model architectures using MACs, assessed spatial complexity using the number of parameters, and measured daily energy consumption on embedded devices such as the NVIDIA Jetson AGX Xavier. Different datasets represent diverse sensor configurations. For HAR models, an increased number of sensors and wearable positions require more parameters and computational resources. The datasets we chose adequately represent various sensor configuration scenarios in HAR, including single and multiple wearable positions. Maintaining low resource consumption, even with a greater number of sensors, is crucial for HAR methods in practical applications.

\begin{figure*}
    \centering
    \includegraphics[width=\textwidth]{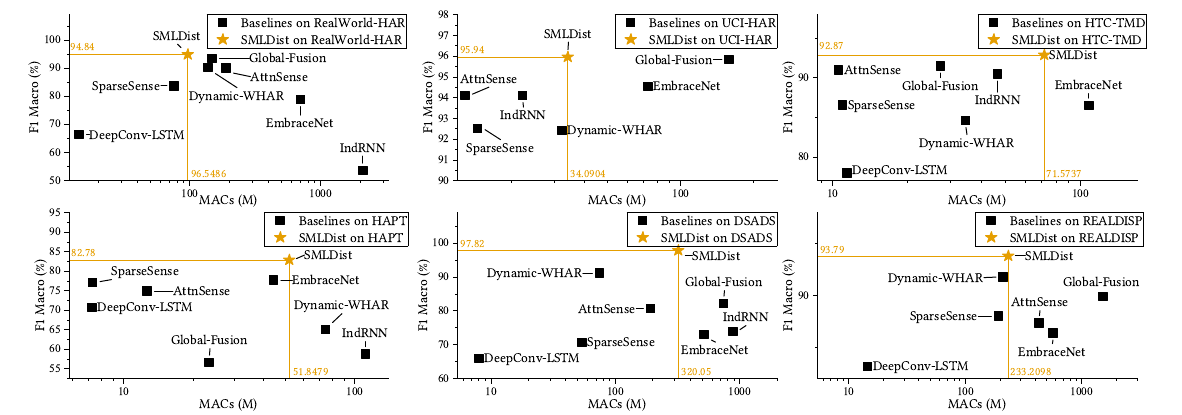}
    \caption{Model performance comparison (F1 macro versus MACs), illustrating the trade-off between HAR recognition accuracy and computational complexity among different model architectures. SMLDist achieves the highest F1 macro at a relatively lower computational complexity.}
    \label{fig:mac-f1-total}
\end{figure*}

\begin{figure*}
    \centering
    \includegraphics[width=\textwidth]{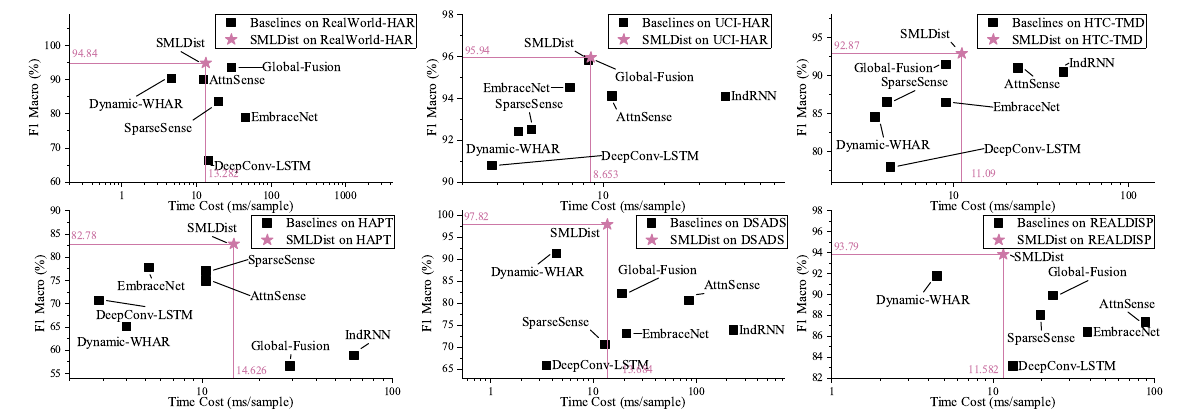}
    \caption{Model performance comparison (F1 macro versus the equivalent time consumption per sample), illustrating the trade-off between HAR recognition accuracy and time expenditure among different model architectures. SMLDist achieves the highest F1 macro at a relatively low time consumption per sample.}
    \label{fig:time-f1-total}
\end{figure*}

\begin{figure*}
    \centering
    \includegraphics[width=\textwidth]{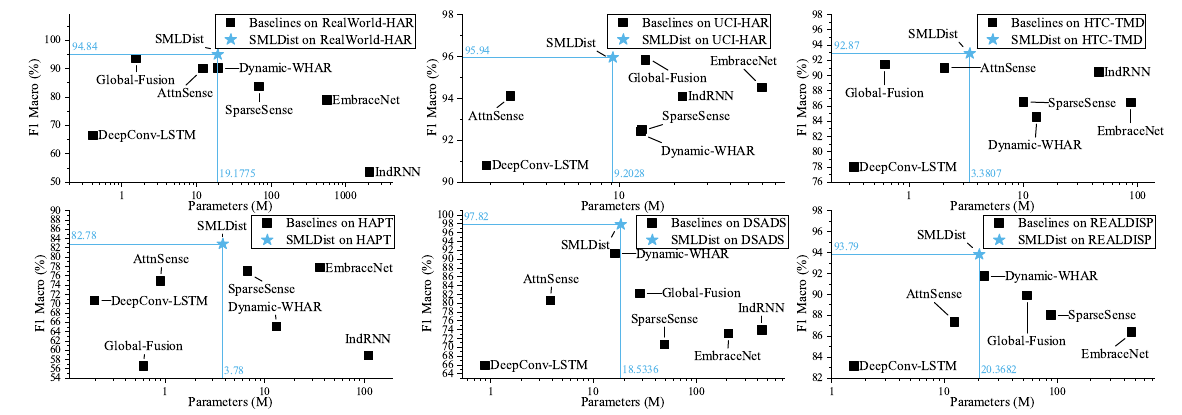}
    \caption{Model performance comparison (F1 macro versus the number of parameters), illustrating the trade-off between HAR recognition accuracy and spatial complexity among different model architectures. SMLDist achieves the highest F1 macro at a relatively low spatial complexity.}
    \label{fig:param-f1-total}
\end{figure*}

In Figure \ref{fig:mac-f1-total}, MACs reflect the computational complexity of the model structure in the corresponding dataset configurations. It is evident in the graph that the SMLDist model maintains a competitively low computational complexity while achieving an excellent balance between computational complexity and classification recognition performance, even when its F1 score significantly outperforms other methods. The computational complexity is directly manifested in the computation time per sample. By comparing the inference times for each sample, Figure \ref{fig:time-f1-total} clearly demonstrates that SMLDist consistently achieves the optimal trade-off between accuracy and computational expense across all scenarios. Compared to computational complexity, SMLDist demonstrates greater efficiency in utilizing spatial complexity as displayed in Figure \ref{fig:param-f1-total}. In achieving optimal classification performance, our SMLDist models rank higher in spatial complexity relative to computational complexity compared to many other model architectures. This indicates that SMLDist achieves a more significant level of parameter compression.

\begin{figure*}
    \centering
    \includegraphics[width=\textwidth]{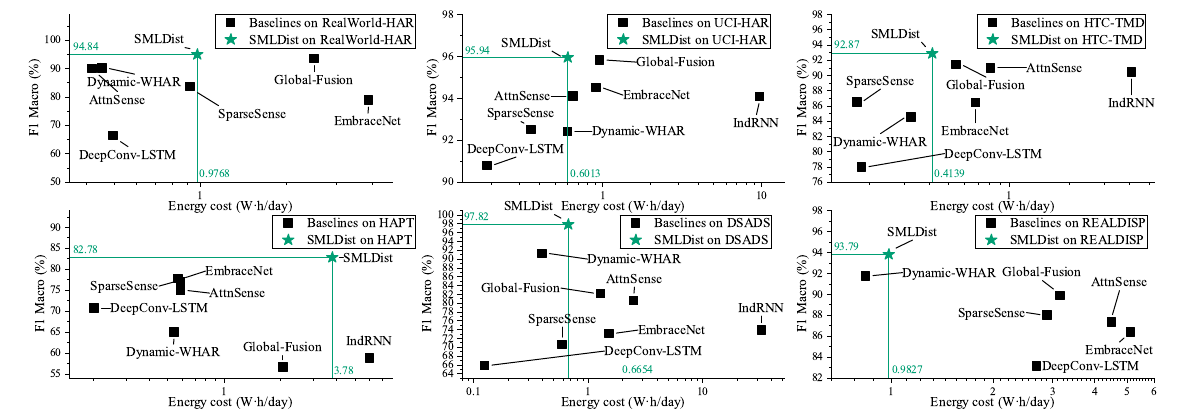}
    \caption{Model performance comparison (F1 macro versus the equivalent daily energy consumption), illustrating the trade-off between HAR recognition accuracy and energy expenditure among different model architectures. SMLDist achieves the highest F1 macro at a relatively low daily energy consumption.}
    \label{fig:energy-f1-total}
\end{figure*}

Building on its demonstrated advantages in computational and spatial complexity, SMLDist further distinguishes itself in energy efficiency, as depicted in Figure \ref{fig:energy-f1-total}. Notably, SMLDist secures the highest F1 macro score while incurring relatively low energy costs, showcasing its effectiveness in HAR recognition with minimal energy expenditure. This attribute underscores SMLDist's economic and environmental benefits, offering a cost-effective solution for continuous operation in real-world applications. Its ability to achieve high accuracy with lower energy consumption aligns with the growing demand for sustainable and efficient AI technologies, emphasizing SMLDist's role in advancing green computing initiatives.

\subsection{Auto-search of Classifiers}

Next, we shift our focus to the model's classifier, which plays a crucial role in making final decisions based on the features obtained from the backbone model and holds significant importance in deep learning models. We conduct a validation analysis of the auto-search mechanism within SMLDist. We will assess the effectiveness of the proposed straightforward mechanism through ablation experiments.

\begin{figure*}
	\centering
	\includegraphics[width=0.75\textwidth]{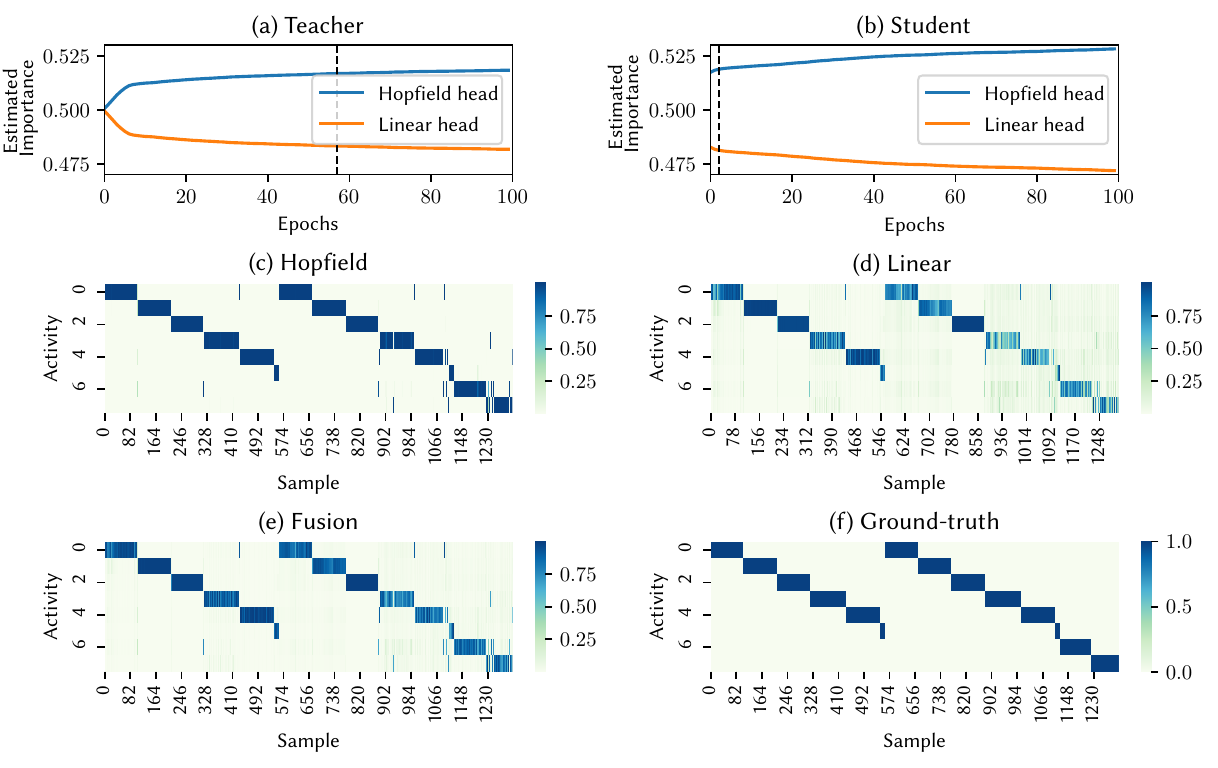}
	\caption{Evaluation of estimated importance $\hat{q}_h$ and predicted probability $\hat{\mathcal{Y}}$ of the model trained on RealWorld-HAR dataset. (a), (b) demonstrate the tendency of $\hat{q}_h$ in SMLDist training process. (c), (d), (e), and (f) demonstrate the sample classification in time order. We can see the importance of the evaluation and selection of classifiers according to (c).}
	\label{fig:simh-evaluation}
\end{figure*}

We assess the estimated importance and predicted probability of both the pre-trained teacher model and its student on the RealWorld dataset. The results are depicted in Figure \ref{fig:simh-evaluation}. To evaluate the auto-searched head, we employ validation sets comprising continuous activities obtained from 2 subjects. Specifically, samples from 0 to 559 are collected from subject 14 in the RealWorld dataset, while the remaining samples are from subject 15. It is important to note that all samples collected from the same subject exhibit temporal continuity.

The classifier's learned $\hat{q}_h$ in the Hopfield network increases, while the weight of the linear classifier decreases. In Figure \ref{fig:simh-evaluation}a, the softmax self-adaptive weights of the Hopfield classifier and the linear classifier during teacher model training are depicted. The dashed vertical line indicates the epoch with the best model performance. Figure \ref{fig:simh-evaluation}b displays the Hopfield classifier's softmax self-adaptive weights and the linear classifier during student model training. Under this condition, the Hopfield classifier's importance continuously decreases, indicating that the stored information in the Hopfield classifier increases while the importance of the linear classifier decreases. The student model inherits the teacher's memory and learning direction. In SMLDist, we fine-tune the student models for only 1 epoch. During the fine-tuning process, the classifier's importance tends to stabilize and undergoes less significant changes.

Utilizing the fusion of heterogeneous classifiers boosts the learning process for the classification task, enabling us to identify the most optimal approach. Figure \ref{fig:simh-evaluation}c illustrates the softmax output logits of all samples in the validation set of the student model's Hopfield classifier. Simultaneously, Figure \ref{fig:simh-evaluation}d presents the corresponding logits from the student model's linear classifier. Upon comparing Figure \ref{fig:simh-evaluation}c to Figure \ref{fig:simh-evaluation}d, it becomes evident that the Hopfield classifier outperforms the linear classifier in accurately predicting a well-trained model. The fused logits, obtained by applying softmax to the output logits of both the Hopfield classifier and the linear classifier, are demonstrated in Figure \ref{fig:simh-evaluation}e. Additionally, Figure \ref{fig:simh-evaluation}f showcases the ground truth activity of the validation set. During the training process, the model's output is determined by the fused logits, which are influenced by both the linear and Hopfield classifiers. In Table \ref{tab:simh-ablation}, we evaluate 4 classifier configurations for ablation evaluation.

In conclusion, classifiers with higher estimated importance probability demonstrate superior performance. It is important to note that a more complex classifier does not necessarily guarantee better performance. Hence, employing an automatic search based on the classifier's importance is a prudent choice.

\begin{table}
	\centering
	\caption{Ablation evaluation of classifier's auto-search. ("H": using only the Hopfield classifier. "L": using only the Hopfield classifier. "A": using the auto-selected classifier. "H+L" using the weighted fusion of the Hopfield and linear classifiers.)}
	\label{tab:simh-ablation}
	\begin{tabular}[width=0.5\textwidth]{c|c|cccc}
		\toprule
		Dataset                        & Metric        & H     & L     & H+L   & A              \\
		\midrule
		\multirow{2}{*}{RealWorld-HAR} & Accuracy (\%) & 84.43 & 89.16 & 95.65 & \textbf{95.73} \\
		                               & F1 Macro (\%) & 77.37 & 85.14 & 94.73 & \textbf{94.84} \\
		\midrule
		\multirow{2}{*}{UCI-HAR}       & Accuracy (\%) & 95.75 & 95.07 & 95.79 & \textbf{96.06} \\
		                               & F1 Macro (\%) & 95.69 & 94.94 & 95.68 & \textbf{95.94} \\
		\midrule
		\multirow{2}{*}{HTC-TMD}       & Accuracy (\%) & 92.98 & 92.85 & 93.10 & \textbf{93.18} \\
		                               & F1 Macro (\%) & 92.71 & 92.56 & 92.76 & \textbf{92.87} \\
		\midrule
		\multirow{2}{*}{HAPT}          & Accuracy (\%) & 88.18 & 90.58 & 90.65 & \textbf{93.02} \\
		                               & F1 Macro (\%) & 72.89 & 72.79 & 74.85 & \textbf{82.78} \\
		\midrule
		\multirow{2}{*}{DSADS}         & Accuracy (\%) & 80.70 & 84.07 & 97.54 & \textbf{97.81} \\
		                               & F1 Macro (\%) & 77.24 & 81.55 & 97.55 & \textbf{97.82} \\
		\midrule
		\multirow{2}{*}{REALDISP}      & Accuracy (\%) & 91.40 & 92.72 & 93.49 & \textbf{94.00} \\
		                               & F1 Macro (\%) & 92.11 & 92.17 & 93.19 & \textbf{93.79} \\
		\bottomrule
	\end{tabular}
\end{table}

\subsection{Discussion of Experiments}

In our experimental evaluation, we extensively deliberate on the optimal energy efficiency ratio attainable by a straightforward lightweight neural network, employing a holistic knowledge distillation pipeline without the necessity for specialized architectural design. SMLDist consistently achieves state-of-the-art performance in human activity recognition (HAR) across a spectrum of challenging scenarios, including user generalization, sensor displacement adaptability, and the discernment of a variety of activity patterns. This performance is validated across eight distinct dataset configurations, substantiating the significant contribution of SMLDist to the precision of HAR.

Assessing the efficacy of knowledge distillation, we demonstrate through a comparative analysis with a suite of contemporary distillation techniques and comprehensive ablation experiments that the more exhaustive knowledge distillation approach embedded within SMLDist positively influences the HAR task. Comparative analyses of HAR models against SMLDist in terms of spatial complexity (number of parameters), temporal complexity (computation time and Multiply-Accumulate operations), and energy expenditure illustrate a marked reduction in resource expenditure across the majority of these metrics. We also delve into the threshold of model compression before performance degradation becomes pronounced, identifying that preserving 60\% of the computational pathways is sufficient to maintain commendable HAR efficacy. Furthermore, we ascertain the optimal classifier search strategy, which entails the determination of the final deployment classifier through the refinement of classifier weights.

\section{Conclusion and Future Work}
\label{section:conclusions-and-future-work}

This paper presents the framework of SMLDist, which is a structural distilling pipeline specifically designed for HAR. SMLDist integrates stage distillation, memory distillation, and logits distillation to construct a multi-level pipeline of knowledge distillation. We demonstrate that cooperative utilization of multiple HAR-specific knowledge sources leads to superior HAR performance compared to relying solely on a single form of knowledge distillation.

Stage distillation is a feature-level knowledge distillation approach that enhances knowledge transfer between models by balancing periodic knowledge and movement tendency knowledge. By introducing the frequency-domain relationship as periodic knowledge, the student model strengthens its perception of the periodic characteristics of HAR samples. Additionally, we have developed an automatic search mechanism that utilizes learnable importance to optimize the classifier for HAR models. This mechanism significantly improves the accuracy of lightweight HAR models. Combining the semantic knowledge from logits distillation with the auto-searched memory knowledge compensates for any limitations in the stage distillation process, resulting in improved robustness for deep HAR models. SMLDist provides an effective method for constructing HAR models based on widely deployed structures like MobileNet. Models optimized by SMLDist achieve reasonable energy costs when deployed on embedded devices. Based on SMLDist, our optimized model has achieved impressive performance in the HAR task.

Despite the significant progress researchers have made in human-centered perception, there are still numerous challenges that remain. Deep learning applications for on-the-go deployments present a wide range of application challenges. Additionally, there are still several limitations in inertia-based fine-grained activity recognition, including behavioral differences among users and the high cost associated with acquiring labeled data. Vision-based intelligent perception continues to face issues related to high model computation and storage overhead. Therefore, exploring the design of new lightweight models and proposing more efficient model compression methods are fascinating research directions.


%

\section*{Acknowledgment}

This work was supported in part by the Strategic Priority Research Program of Chinese Academy of Sciences under Grant XDA28040500, the National Natural Science Foundation of China under Grant 62261042, the Key Research Projects of the Joint Research Fund for Beijing Natural Science Foundation and the Fengtai Rail Transit Frontier Research Joint Fund under Grant L221003, and the Beijing Natural Science Foundation under Grant 4232035 and 4222034.




\bibliographystyle{IEEEtran}
\bibliography{sample-base}
%



%

\begin{IEEEbiography}[{\includegraphics[width=1in,height=1.25in,clip,keepaspectratio]{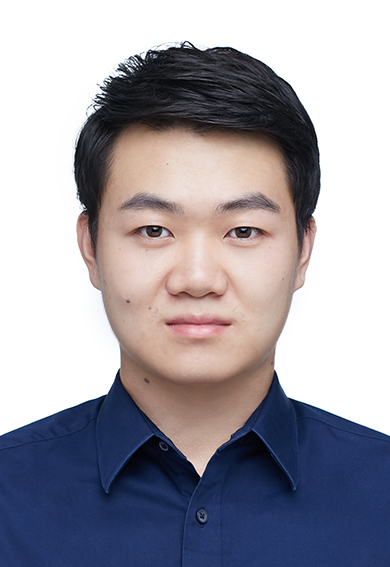}}]{Runze Chen}
	received the B.S. degree from the School of Software Engineering, Beijing University of Posts and Telecommunications, Beijing, China, in 2019. He is working toward the Ph.D. degree in Beijing University of Posts and Telecommunications and is a visiting student in the Institute of Computer Technology, Chinese Academy. His research interests include: mobile computing, autonomous vehicles and mobile intelligence.
\end{IEEEbiography}

\begin{IEEEbiography}[{\includegraphics[width=1in,height=1.25in,clip,keepaspectratio]{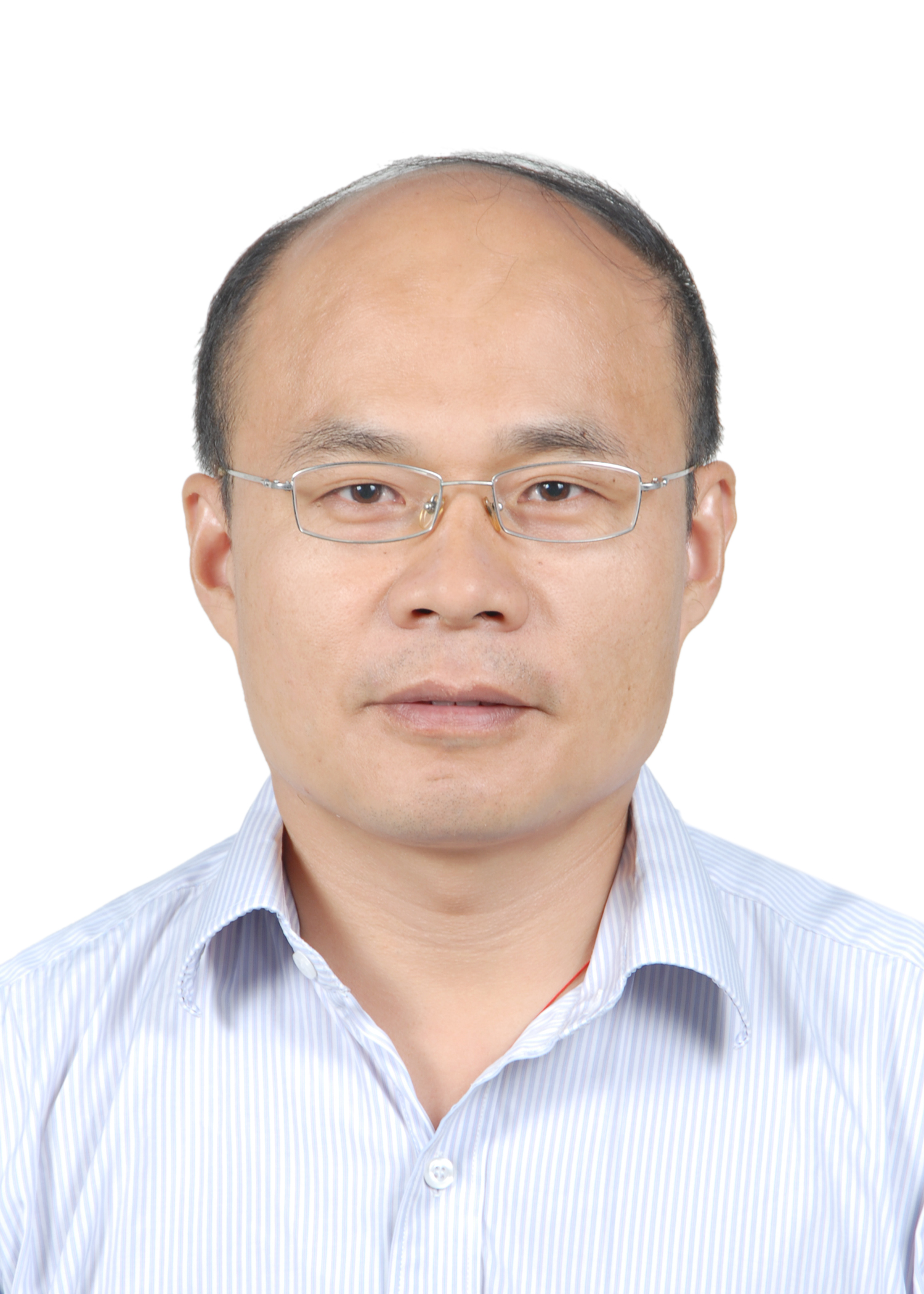}}]{Haiyong Luo}
	received the B.S. degree from the Department of Electronics and Information Engineering, Huazhong University of Science and Technology, Wuhan, China, in 1989, the M.S. degree from the School of Information and Communication Engineering, Beijing University of Posts and Telecommunication, China, in 2002, and the Ph.D. degree in computer science from the University of Chines Academy of Sciences, Beijing, China, in 2008. He is currently an Associate Professor with the Institute of Computer Technology, Chinese Academy of Science, China. His main research interests are location-based services, pervasive computing, mobile computing, and Internet of Things. He is a member of the IEEE.
\end{IEEEbiography}


\begin{IEEEbiography}[{\includegraphics[width=1in,height=1.25in,clip,keepaspectratio]{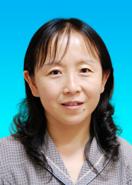}}]{Fang Zhao}
	received the B.S. degree from the School of Computer Science and Technology, Huazhong University of Science and Technology, Wuhan, China, in 1990, the M.S. and Ph.D. degrees in computer science and technology from the Beijing University of Posts and Telecommunications, Beijing, China, in 2004 and 2009, respectively. She is currently a Professor with the School of Software Engineering, Beijing University of Posts and Telecommunication. Her research interests include mobile computing, location-based services, and computer networks.
\end{IEEEbiography}

\begin{IEEEbiography}[{\includegraphics[width=1in,height=1.25in,clip,keepaspectratio]{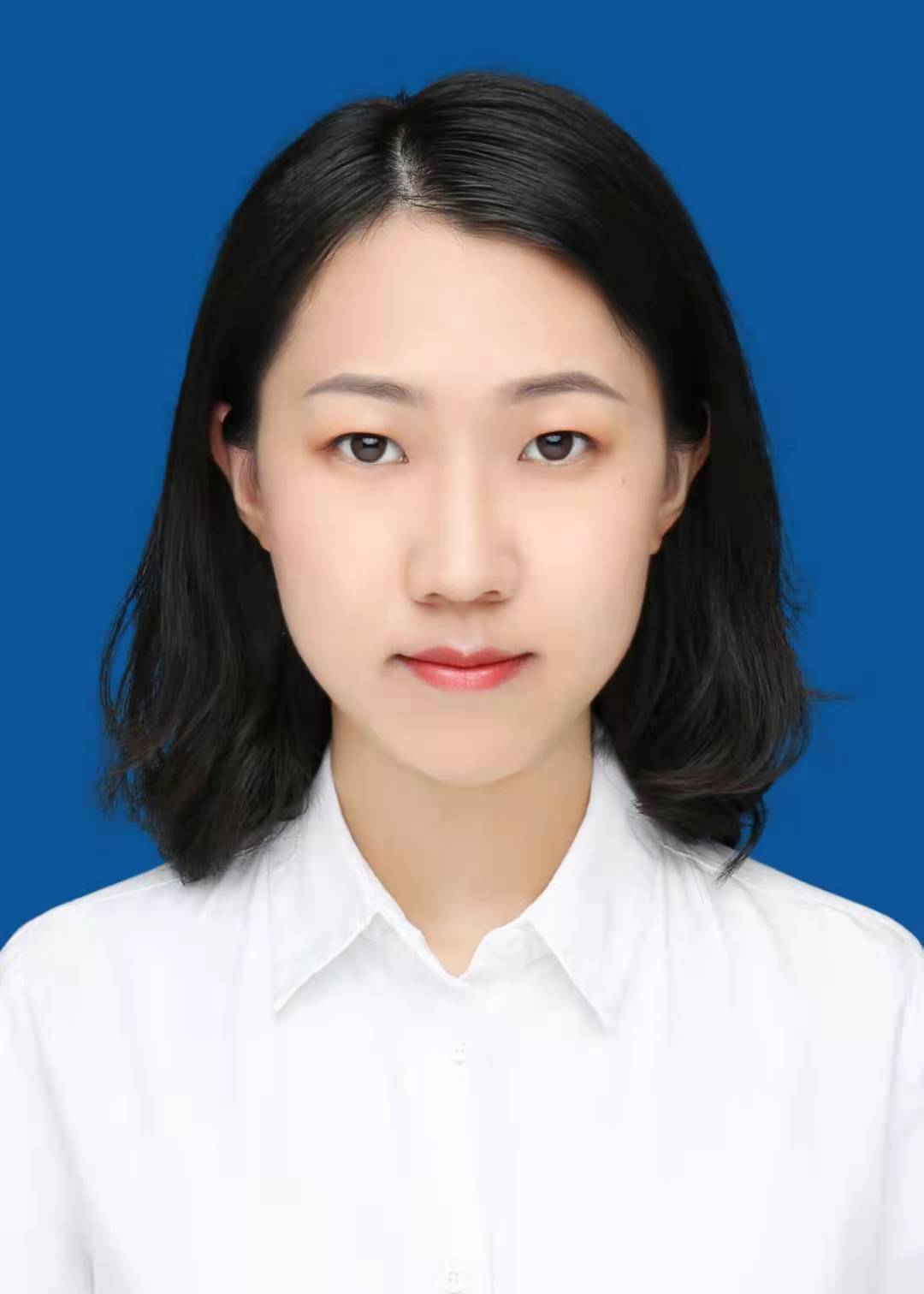}}]{Xuechun Meng} received the B.S. degree in Computer Science and Technology from the School of Computer Science and Technology at China University of Mining and Technology, Xuzhou, Jiangsu, China, in 2020. She obtained her M.Eng. degree in Software Engineering from the School of Computer Science and Technology at Beijing University of Posts and Telecommunications, Beijing, China, in 2023. During her involvement in this paper, she was studying at Beijing University of Posts and Telecommunications and was a visiting student at the Institute of Computer Technology, Chinese Academy of Sciences. Her interests include simultaneous localization and mapping, and semantic segmentation.

\end{IEEEbiography}

\begin{IEEEbiography}[{\includegraphics[width=1in,height=1.25in,clip,keepaspectratio]{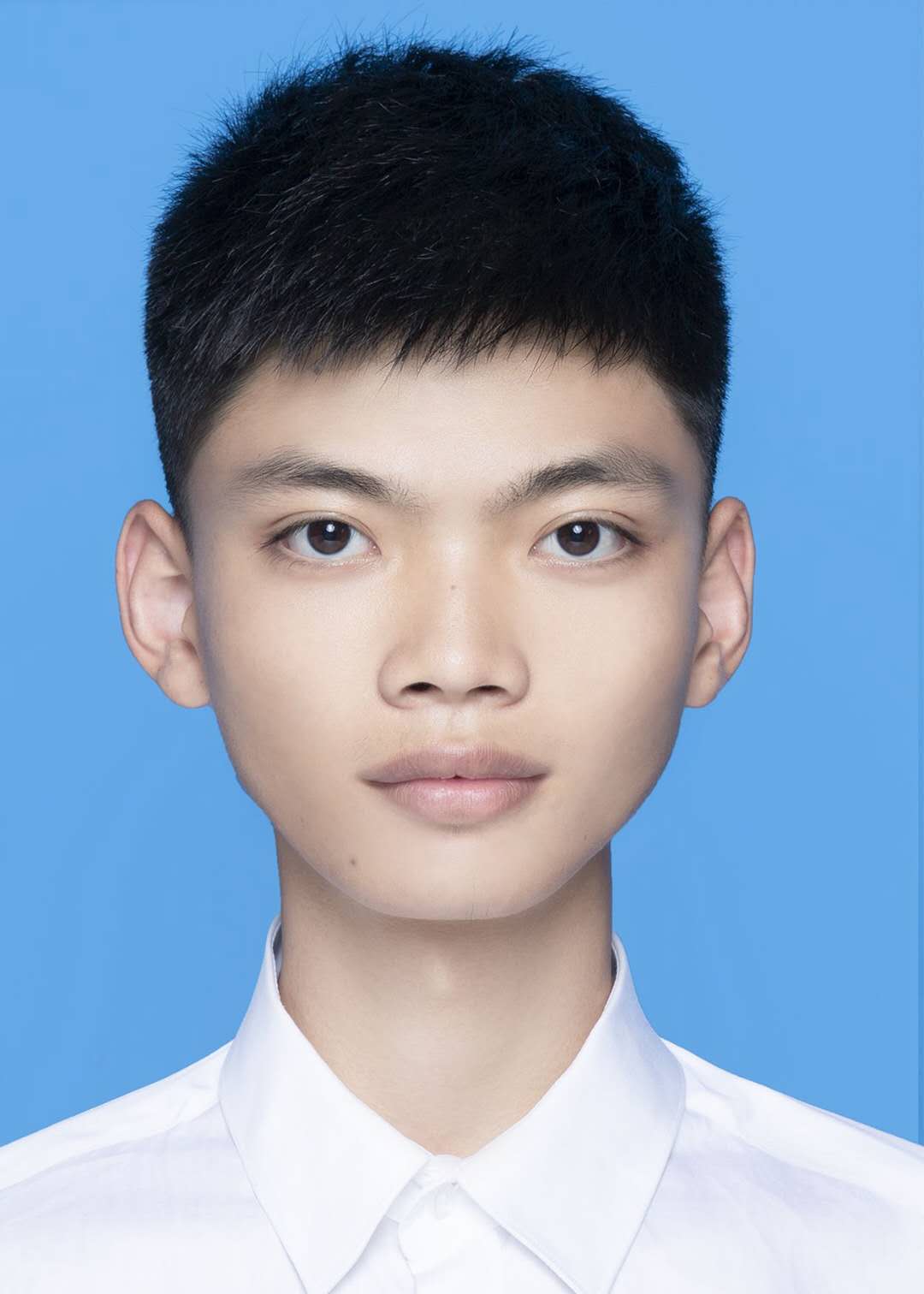}}]{Zhiqing Xie} received the B.S. degree in Computer Science and Technology from the School of Computer Science and Technology at China University of Mining and Technology, Xuzhou, Jiangsu, China, in 2020. He obtained his M.Eng. degree in Software Engineering from the School of Computer Science and Technology at Beijing University of Posts and Telecommunications, Beijing, China, in 2023. During his involvement in this paper, he was studying at Beijing University of Posts and Telecommunications and was a visiting student at the Institute of Computer Technology, Chinese Academy of Sciences. His research interests include: mobile computing and mobile intelligence.
\end{IEEEbiography}

\begin{IEEEbiography}[{\includegraphics[width=1in,height=1.25in,clip,keepaspectratio]{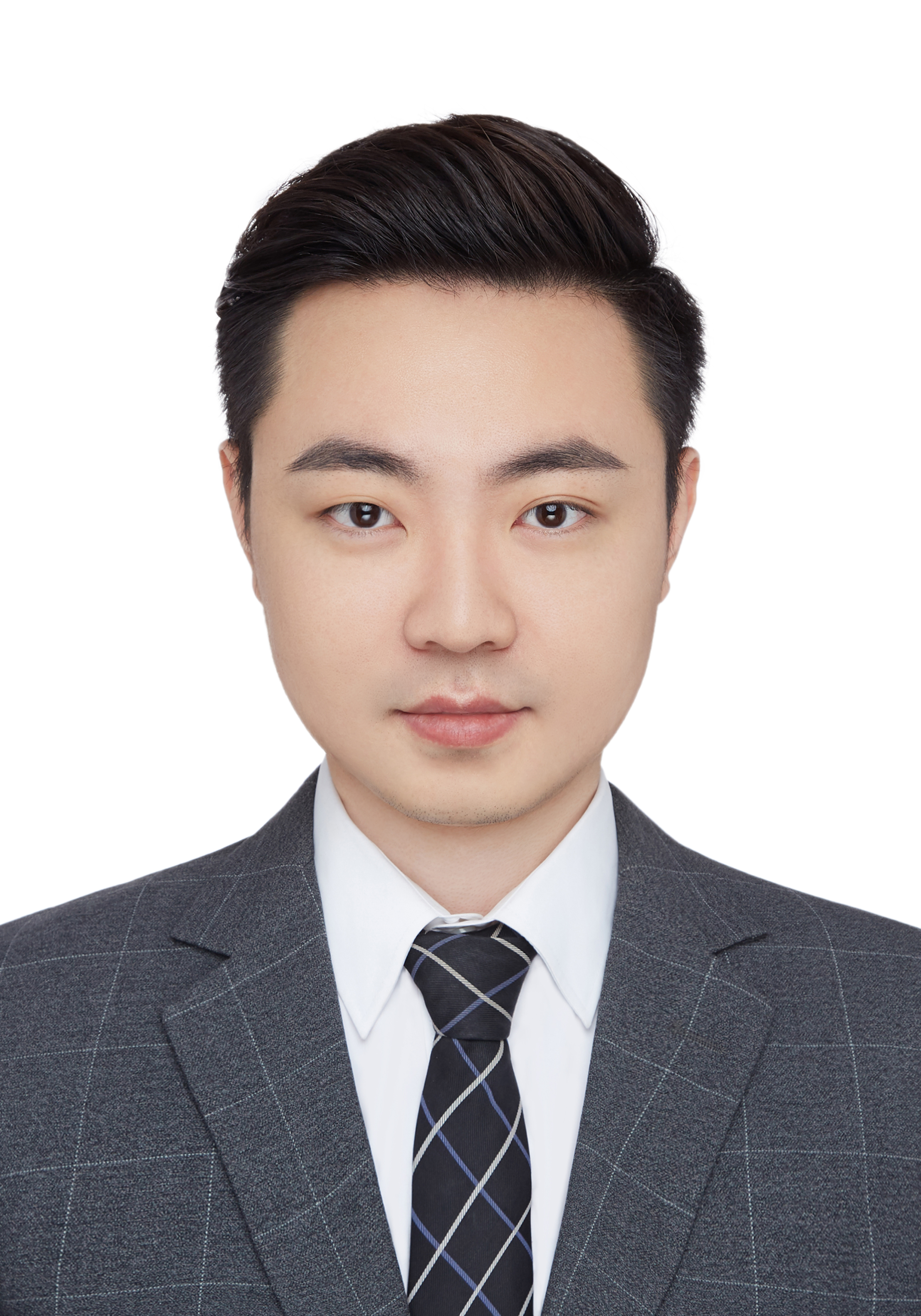}}]{Yida Zhu} received the B.S. degree in Software Engineering from the School of Software Engineering at Beijing University of Posts and Telecommunications, Beijing, China, in 2017. He obtained his Ph.D. degree in Software Engineering from the same institution in 2022. During his involvement in this paper, he was a Ph.D. candidate at the School of Software Engineering, Beijing University of Posts and Telecommunications, and was a visiting student at the Institute of Computer Technology, Chinese Academy. He is currently working at Meituan. His current main interests include location-based services, pervasive computing, deep learning, transfer learning, and machine learning.

\end{IEEEbiography}




\end{document}